\documentclass{article}

\usepackage{arxiv}

\usepackage[hyphens]{url}
\usepackage{graphicx}
\usepackage{caption}

\usepackage{amsmath,amssymb,amsthm}
\usepackage{xspace}
\usepackage{epsfig}
\usepackage{makecell}

\usepackage{algorithm}
\usepackage[noend]{algorithmic}
\usepackage{color}
\usepackage{url}
\usepackage{booktabs,multirow}
\usepackage{hyperref}

\newtheorem{theorem}{Theorem}

\newtheorem{lemma}[theorem]{Lemma}

\newtheorem{definition}{Definition}

\renewcommand{\epsilon}{\varepsilon}

\newcommand{\E}[1]{\text{E}\left(#1\right)}

\newcommand{\Prob}[1]{\text{Pr}\left(#1\right)}

\newcommand{\RLS}{\text{RLS}\xspace}
\newcommand{\EA}{\text{(1+1)~EA}\xspace}
\newcommand{\MA}{ILS\xspace}

\newcommand{\wlo}{w.\,l.\,o.\,g.\xspace}

\DeclareMathOperator{\neighborhood}{\mathcal{N}}

\newcommand{\col}{c}

\newcommand{\chromatic}{\chi}

\definecolor{black}{rgb}{0.0,0.0,0.0}
\definecolor{grey}{rgb}{0.3,0.3,0.3}

\usepackage{tikz}
\usepackage{verbatim}
\usetikzlibrary{arrows,shapes}
\usetikzlibrary{calc}

\tikzstyle{color1}=[fill=red!40!white]
\tikzstyle{color2}=[fill=blue!40!white]
\tikzstyle{color3}=[fill=green!50!white]
\tikzstyle{color4}=[fill=orange!50!white]
\tikzstyle{color5}=[fill=violet!30!white]
\tikzstyle{color6}=[fill=yellow!30!white]

\newcommand{\thmref}[1]{{\footnotesize{}\color{gray!60!black}[Thm~\ref{#1}]}}

\newcommand{\aware}{tailored\xspace}
\newcommand{\Aware}{Tailored\xspace}
\newcommand{\unaware}{generic\xspace}

\begin{document}

\title{Time Complexity Analysis of Randomized Search Heuristics for the Dynamic Graph Coloring Problem}

\author{
   Jakob Bossek \\
   Statistics and Optimization \\
   University of M\"unster \\
   M\"unster, Germany \\
   \texttt{bossek@wi.uni-muenster.de\,\,\,}
   \And
   Frank Neumann \\
   School of Computer Science \\
   The University of Adelaide \\
   Adelaide, Australia \\
   \texttt{frank.neumann@adelaide.edu.au}
   \And
   Pan Peng \\
   Department of Computer Science \\
   University of Sheffield \\
   Sheffield, United Kingdom \\
   \texttt{\,\,\,\,\,\,\,\,\,p.peng@sheffield.ac.uk\,\,\,\,\,\,\,\,\,}
   \And
   Dirk Sudholt \\
   Chair of Algorithms for Intelligent Systems \\
   University of Passau \\
   Passau, Germany \\
   \texttt{Dirk.Sudholt@uni-passau.de}
} 

\maketitle

\begin{abstract}
We contribute to the theoretical understanding of randomized search heuristics for dynamic problems. We consider the classical vertex coloring problem on graphs and investigate the dynamic setting where edges are added to the current graph. We then analyze the expected time for randomized search heuristics to recompute high quality solutions. The (1+1)~Evolutionary Algorithm and RLS operate in a setting where the number of colors is bounded and we are minimizing the number of conflicts. Iterated local search algorithms use an unbounded color palette and aim to use the smallest colors and, consequently, the smallest number of colors.

We identify classes of bipartite graphs where reoptimization is as hard as or even harder than optimization from scratch, i.e., starting with a random initialization. Even adding a single edge can lead to hard symmetry problems. However, graph classes that are hard for one algorithm turn out to be easy for others. In most cases our bounds show that reoptimization is faster than optimizing from scratch.
We further show that tailoring mutation operators to parts of the graph where changes have occurred can significantly reduce the expected reoptimization time. In most settings the expected reoptimization time for such tailored algorithms is linear in the number of added edges. However, tailored algorithms cannot prevent exponential times in settings where the original algorithm is inefficient.

\end{abstract}

\keywords{Evolutionary algorithms \and dynamic optimization \and running time analysis}

\section{Introduction}
Evolutionary algorithms (EAs) and other bio-inspired computing techniques have been used for a wide range of complex optimization problems~\cite{DBLP:books/daglib/0034477,DBLP:books/sp/chiong12}. They are easy to apply to a newly given problem and are able to adapt to changing environments. This makes them well suited for dealing with dynamic problems where components of the given problem change over time~\cite{NGUYEN20121,DBLP:series/isrl/RichterY13}. 

We contribute to the theoretical understanding of evolutionary algorithms in dynamically changing environments. Providing a sound theoretical basis on the behaviour of these algorithms in changing environments helps to develop better performing algorithms through a deeper understanding of their working principles.

Dynamic problems have been studied in the area of runtime analysis for simple algorithms such as randomized local search (\RLS) and the classical \EA. An overview on rigorous runtime results for bio-inspired computing techniques in stochastic and dynamic environments can be found in \cite{DBLP:journals/corr/abs-1806-08547}.
Early work focused on artificial problems like a dynamic \textsc{OneMax} problem~\cite{DrosteDynamic}, the function \textsc{Balance}~\cite{RohlfshagenLehreYao2009} where rapid changes can be beneficial, the function \textsc{MAZE} that features an oscillating behavior~\cite{DBLP:conf/ppsn/KotzingM12} and problems involving
moving Hamming balls~\cite{Dang2017}.
The investigations of the \EA for a dynamic variant of the classical \textsc{LeadingOnes} problem in \cite{DBLP:conf/gecco/DoerrD019} reveal that previous optimization progress might (almost) be completely lost even if small perturbations of the problem occur. This motivated the introduction of a population-based structural diversity optimization approach~\cite{DBLP:conf/gecco/DoerrD019}. The approach is able to maintain structural progress by preserving solutions of beneficial structure although they might have low fitness after a dynamic change has occurred.

In terms of classical combinatorial optimization problems, prominent problems such as single-source-shortest-paths~ \cite{DBLP:journals/tcs/LissovoiW15}, makespan scheduling~\cite{DBLP:conf/ijcai/NeumannW15}, and the vertex cover problem~\cite{DBLP:conf/gecco/PourhassanGN15,DBLP:conf/ssci/PourhassanRN17,DBLP:journals/algorithmica/ShiNW21} have been investigated in a dynamic setting. Furthermore, the behaviour of evolutionary algorithms on linear functions with dynamic constraints has been analyzed in~\cite{DBLP:journals/algorithmica/ShiSFKN19,DBLP:journals/algorithmica/ShiSFKN20} and experimental investigations for the knapsack problem with a dynamically changing constraint bound have been carried out in~\cite{DBLP:conf/ppsn/Roostapour0N18}.
These studies have been extended in \cite{DBLP:journals/corr/abs-1811-07806} to a broad class of problems and the performance of an evolutionary multi-objective algorithm has been analyzed in terms of its approximation behaviour dependent on the submodularity ratio of the considered problem.

We consider graph vertex coloring, a classical NP-hard optimization problem. In the context of problem specific approaches, algorithms have been designed to update solutions after a dynamic change has happened. Dynamic algorithms have been proposed to maintain proper coloring for graphs with maximum degree at most $\Delta$,\footnote{In such graphs, there always exist a proper $(\Delta+1)$-vertex coloring. Furthermore, such a proper coloring can be found in linear time.} with the goal of using as few colors as possible while keeping the (amortized) update time small~\cite{barenboim2017fully,BCHN18:coloring}. There exist algorithms that aim to perform as few (amortized) vertex recolorings as possible in order to maintain a proper coloring in a dynamic graph~\cite{barba2017dynamic,SW18:coloring}. There have also been studies of \emph{$k$-list coloring} in a dynamic graph such that each update corresponds to adding one vertex (together with the incident edges) to the graph (e.g.~\cite{Hartung2013}). The related problem of maintaining a coloring with minimal total colors in a \emph{temporal} graph has recently been studied \cite{DBLP:journals/corr/abs-1811-04753}. From a practical perspective, incremental algorithms or heuristics have been proposed that update the graph coloring by exploring a small number of vertices~\cite{Preuveneers2004acodygra,Yuan2017}.


Graph coloring has been studied for specific local search and evolutionary algorithms in~\cite{Fischer2005,Sudholt2005,Sudholt2010b}.
\cite{Fischer2005} studied a problem inspired by the Ising model in physics that on bipartite graphs is equivalent to the vertex coloring problem. They showed that on cycle graphs the \EA and RLS find optimal colorings in expected time $O(n^3)$. This bound is tight under a sensible assumption. They also showed that crossover can speed up the optimization time by a factor of~$n$. \cite{Sudholt2005} showed that on complete binary trees the \EA needs exponential expected time, whereas a Genetic Algorithm with crossover and fitness sharing finds a global optimum in $O(n^3)$ expected time. \cite{Sudholt2010b} considered a different representation with unbounded-size palettes, where the goal is to use small color values as much as possible. They considered \emph{iterated local search (ILS)} algorithms with operators based on so-called \emph{Kempe chains} that are able to recolor large connected parts of the graph, while maintaining feasibility. This approach was shown to be efficient on paths and for coloring planar graphs of bounded degree $(\Delta \le 6)$ with 5 colors. The authors also gave a worst-case graph, a tree, where Kempe chains fail, but a new operator called \emph{color elimination} that performs Kempe chains in parallel, succeeds in 2-coloring all bipartite graphs efficiently. Table~\ref{tab:all-times} (top rows) gives an overview over previous results.

We revisit these algorithms and graph classes for a dynamic version of the vertex coloring problem. We assume that the graph is altered by adding up to $T$ edges to it. This may create new conflicts that need to be resolved. Note that deleting edges from the graph can never worsen the current coloring, hence we focus on adding edges only.\footnote{In general, the chromatic number of a graph can decrease when removing edges. We focus on graphs that can be colored with 2 or 5 colors, respectively. For 2-colorable graphs the chromatic number can only decrease if the graph becomes empty. For our results on 5-coloring graphs the true chromatic number will be irrelevant.} The assumption that the graph is updated by adding edges is natural in many practical scenarios. For example, the web graph is explored gradually by a crawler that adds edges as they are discovered; the citation networks (in which nodes are research papers and edges indicate the citations between two papers) and collaboration networks (in which nodes are scientific researchers and edges correspond to collaborations) grow by adding edges.
Our goal is to determine the expected reoptimization time, that is, the time to rediscover a proper coloring after up to $T$ edges have been added, {given that the previous graph is properly colored,} and how this time depends on~$T$ and the number of vertices~$n$. Our results are summarized in Table~\ref{tab:all-times} (center row in each of the two tables).

We start by considering bipartite graphs in Section~\ref{sec3}. We find that even adding a single edge can create a hard symmetry problem for RLS and the \EA: expected reoptimization times for paths and binary trees are as bad as, or even slightly worse, than the corresponding bounds for optimizing from scratch, i.e.\ starting with a random initialization. In contrast, ILS with Kempe chains or color elimination reoptimizes these instances efficiently. While ILS with color eliminations reoptimizes every bipartite graph in expected time $O(\sqrt{T}n \log n)$ or better, ILS with Kempe chains needs expected time $\Theta(2^{n/2})$ even when connecting a tree with an isolated edge. This instance is easy for all other algorithms as they all have reoptimization time $O(n \log^+ T)$ (where $\log^+ T = \max\{1, \log T\}$ is used to avoid expressions involving a factor of $\log T$ becoming~0 when $T=1$).

In Section~\ref{sec4} we show that ILS with either operator is also able to efficiently rediscover a 5-coloring for planar graphs with maximum degree $\Delta \le 6$ in expected time $O(n \log^+ T)$.

In Section~\ref{sec5} we design mutation operators that focus on the areas in the graph where a dynamic change has happened. We call these algorithms \emph{\aware} algorithms and refer to the original algorithms as \emph{\unaware} algorithms. We show that \aware algorithms can reoptimize most graph classes in time $O(T)$ after inserting $T$ edges, however they cannot prevent exponential times in cases where the corresponding generic algorithm is inefficient. All our results are shown in Table~\ref{tab:all-times} (bottom rows).


Section~\ref{sec2} defines the considered algorithms and the setting of reoptimization. It briefly reviews the computational complexity of executing one iteration of each algorithm as well as related work on problem-specific algorithms.

A preliminary version with parts of the results was published in~\cite{BNPS2019}. While results for \aware algorithms were limited to adding one edge, results in this extension hold for adding up to $T$ edges. This required a major redesign of the \aware algorithms and entirely new proofs for some graph classes. We also added a new structural insight on ILS: Lemma~\ref{lem:number-of-Delta-and-Delta-plus-1-non-increasing} establishes that the number of vertices colored with one of the two largest possible colors, $\Delta+1$ and $\Delta$, cannot increase over time. This simplifies several analyses, improves our previous upper bound for ILS on binary trees from $O(n \log n)$ to $O(n \log^+ T)$, and generalises the latter result to larger classes of graphs (see Theorem~\ref{the:general-n-log-T-bound-for-ils-with-ce}).
We also improved our exponential lower bound for the generic and tailored \EA on binary trees by a factor of~$n$ and added a tight upper bound (see Theorems~\ref{theorem:tree} and~\ref{theorem:aware-lower-tree}).

\begin{table}[btp]
\caption{Worst-case expected times of tailored and generic algorithms for bounded-size palettes (top) and unbounded-size palettes (bottom) for (re-)discovering proper 2-colorings for bipartite graphs and proper 5-colorings for planar graphs. In the dynamic setting, up to $T$ edges are added to the graph. We use the notation $\log^+ T = \max\{1, \log T\}$. The upper bounds for ILS with color eliminations on general bipartite graphs improve to $O(n \log^+ T)$ for \unaware ILS and $O(T)$ for \aware ILS if no end point of an added edge is neighbored to an end point of another added edge, of if $\Gamma \le 4$.
}
\label{tab:all-times}
\setlength{\tabcolsep}{13pt}
\renewcommand{\arraystretch}{1.20}
\centering
\begin{tabular}{p{2.0cm}lp{4cm}p{4.5cm}}
{\bf Setting} & {\bf Graph class} & {\bf \EA} & {\bf RLS}\\
\toprule
\multirow{4}{*}{Static} & paths & $O(n^3)$~\cite{Fischer2005} & $O(n^3)$~\cite{Fischer2005}\\
& binary trees & $\exp(\Omega(n))$~\cite{Sudholt2005} & $\infty$~\thmref{theorem:tree} \\
& depth-2 star & $O(n \log n)$ \thmref{the:rls-on-worst-case-tree-static} & $O(n \log n)$ \thmref{the:rls-on-worst-case-tree-static}\\
& any bipartite & $\exp(\Omega(n))$~\cite{Sudholt2005} & $\infty$~\thmref{theorem:tree}\\
\midrule
\multirow{4}{*}{\shortstack[l]{Dynamic\\(\unaware\\algorithms)}} & paths & $\Theta(n^3)$ \thmref{theorem:path} & $\Theta(n^3)$ \thmref{theorem:path}\\
& binary trees & $\Theta(n^{(n+1)/4})$ \thmref{theorem:tree} & $\infty$~\thmref{theorem:tree}\\
& depth-2 star & $O(n \log^+ T)$ \thmref{the:rls-on-worst-case-tree-static} & $O(n \log^+ T)$ \thmref{the:rls-on-worst-case-tree-static}\\
& any bipartite & $\Omega(n^{(n+1)/4})$ \thmref{theorem:tree} & $\infty$~\thmref{theorem:tree}\\
\midrule
\multirow{4}{*}{\shortstack[l]{Dynamic\\(\aware\\algorithms)}} & paths & $O(n^2)$~\thmref{the:multi-aware-path-ea-rls} & $O(n^2 \log^+ T)$~\thmref{the:multi-aware-path-ea-rls} \\
& binary trees & $\Theta(n^{(n-3)/4})$~\thmref{theorem:aware-lower-tree} & $\infty$~\thmref{theorem:aware-lower-tree}\\
& depth-2 star & $O(\log^+ T)$~\thmref{the:multi-aware-rls-on-worst-case-tree-static} & $O(T)$~\thmref{the:multi-aware-rls-on-worst-case-tree-static}  \\
& any bipartite & $\Omega(n^{(n-3)/4})$~\thmref{theorem:aware-lower-tree} & $\infty$~\thmref{theorem:aware-lower-tree}\\
\bottomrule
\end{tabular}
\vskip11pt
\begin{tabular}{p{2.0cm}lp{4cm}p{4.5cm}}
{\bf Setting} & {\bf Graph class} & {\bf ILS$+$Kempe Chains\!\!\!} & {\bf ILS$+$Color Eliminations}\\
\toprule
\multirow{5}{*}{Static} & paths & $O(n)$~\cite{Sudholt2010b} & $O(n \log n)$~\thmref{the:ils-on-paths}\\
& binary trees & $O(n \log n)$~\thmref{the:ils-on-trees} & $O(n \log n)$~\thmref{the:ils-on-trees}\\
& depth-2 star & $\exp(\Omega(n))$~\cite{Sudholt2010b} & $O(n^2 \log n)$~\cite{Sudholt2010b}\\
& any bipartite & $\exp(\Omega(n))$~\cite{Sudholt2010b} & $O(n^2 \log n)$~\cite{Sudholt2010b}\\
& planar, $\Delta \le 6$ & $O(n \log n)$~\cite{Sudholt2010b} & $O(n \log n)$~\thmref{thm:planar}\\
\midrule
\multirow{5}{*}{\shortstack[l]{Dynamic\\(\unaware\\algorithms)}} & paths & $O(n)$~\cite{Sudholt2010b} & $O(n \log^+ T)$ \thmref{the:ils-on-paths}\\
& binary trees & $O(n \log^+ T)$~\thmref{the:ils-on-trees} & $O(n \log^+ T)$ \thmref{the:ils-on-trees}\\
& depth-2 star & $\Theta(2^{n/2})$ \thmref{the:worst-case-for-Kempe-chains} & $O(n \log^+ T)$ \thmref{the:color-eliminations-on-worst-case-tree}\\
& any bipartite & $\Omega(2^{n/2})$~\thmref{the:worst-case-for-Kempe-chains} & $O(\min\{\sqrt{T}, \Gamma\} n \log n)$~\thmref{the:ils-ce-rediscover-time}\\
& planar, $\Delta \le 6$ & $O(n \log^+ T)$ \thmref{thm:planar} & $O(n \log^+ T)$ \thmref{thm:planar}\\
\midrule
\multirow{5}{*}{\shortstack[l]{Dynamic\\(\aware\\algorithms)}} & paths & $O(T)$~\thmref{the:aware-T-path-tree-Kempe} & $O(T)$~\thmref{the:aware-T-bipartite-color-elimination} \\
& binary trees & $O(T)$~\thmref{the:aware-T-path-tree-Kempe} & $O(T)$~\thmref{the:aware-T-bipartite-color-elimination} \\
& depth-2 star & $\Theta(2^{n/2})$~\thmref{the:aware-worst-case-tree} & $O(T)$~\thmref{the:aware-T-bipartite-color-elimination} \\
& any bipartite & $\Omega(2^{n/2})$~\thmref{the:aware-worst-case-tree} & $O(\min\{\sqrt{T}, \Gamma\}n)$~\thmref{the:aware-T-bipartite-color-elimination} \\
& planar, $\Delta \le 6$ & $O(T)$~\thmref{the:aware-T-planar} & $O(T)$~\thmref{the:aware-T-planar} \\
\bottomrule
\end{tabular}
\end{table}

\section{Preliminaries}
\label{sec2}
Let $G = (V, E)$ denote an undirected graph with vertices~$V$ and
edges~$E$. We denote by $n := |V|$ the number of vertices in $G$. {We let $\Delta$ denote the maximum degree of the graph $G$.}
A \emph{vertex coloring} of $G$ is an assignment $c : V \to \{1, \ldots, n\}$ of color values to the vertices of $G$.
Let $\deg(v)$ be the degree of a vertex~$v$ and $c(v)$ be its color in the current coloring.
Every edge $\{u, v\} \in E$ where $c(v) = c(u)$ is called a \emph{conflict}. A color is called \emph{free} for a vertex $v \in V$ if it is not assigned to any neighbor of $v$. The chromatic number $\chromatic(G)$ is the minimum number of colors that allows for a conflict-free coloring. A coloring is called \emph{proper} is there is no conflicting edge.


\subsection{Algorithms with Bounded-Size Palette}
In this representation, the total number of colors is fixed, i.e., the color palette has fixed size $k\leq n$. The search space is $\{1, \dots, k\}^n$ and the objective function is to \emph{minimize the number of conflicts}.

We assume that in the static setting all algorithms are initialized uniformly at random. In a dynamic setting we assume that a proper $k$-coloring $x$ has been found. Then the graph is changed dynamically and $x$ becomes an initial solution for the considered algorithms.

We define the dynamic \EA for this search space as follows (see Algorithm~\ref{alg:oneplusone}). Assume that the current solution is $x$. We consider all algorithms as infinite processes as we are mainly interested in the expected number of iterations until good solutions are found or rediscovered.
\begin{algorithm}[htb]
    \caption{\EA ($x$)}
    \algsetup{indent=1.5em}
    \begin{algorithmic}[1]
        \WHILE{optimum not found}
        \STATE Generate $y$ by deciding to mutate each component $x_i$ with probability $1/n$: if yes, choose a new value $y_i \in \{1, \dots, k\} \setminus \{x_i\}$ uniformly at random.
        \STATE If $y$ has no more conflicts than $x$, let $x := y$.
        \ENDWHILE
    \end{algorithmic}
    \label{alg:oneplusone}
\end{algorithm}

We also define randomized local search (RLS; see Algorithm~\ref{alg:rls}) as a variant of the \EA where exactly one component is mutated.
\begin{algorithm}[htb]
    \caption{RLS ($x$)}
    \algsetup{indent=1.5em}
    \begin{algorithmic}[1]
        \WHILE{optimum not found}
        \STATE Generate $y$ by choosing an index $i \in \{1, \dots, n\}$ uniformly at random, choosing a new value $y_i \in \{1, \dots, k\} \setminus \{x_i\}$ uniformly at random and setting $y_j = x_j$ for all $j \neq i$.
        \STATE If $y$ has no more conflicts than $x$, let $x := y$.
        \ENDWHILE
    \end{algorithmic}
    \label{alg:rls}
\end{algorithm}

\subsection{Algorithms with Unbounded-Size Palette}
\label{sec:algorithms}

In this representation, the color palette size is sufficiently large (say has size~$n$). Our goal is to maintain a \emph{proper} vertex coloring and to reward colorings that color many vertices with small color values. The motivation for focusing on small color values is to introduce a direction for the search process to use a small set of preferred colors and the hope is that large color values eventually become obsolete.
We use the selection operator from~\cite[Definition~1]{Sudholt2010b} that defines a color-occurrence vector counting the number of vertices colored with given colors and tries to evolve a color-occurrence vector that is as close to optimum as possible.
\begin{definition}\label{def:color_occurrence}\cite{Sudholt2010b}
For $x, y$ we say that $x$ is better than $y$ and write $x \succeq y$ iff\vspace*{-1.0ex}
\begin{itemize}
 \item $x$ has fewer conflicting edges than~$y$ or
 \item $x$ and $y$ have an equal number of conflicting edges and their color frequencies are lexicographically ordered as follows. Let $n_i(x)$ be the number of $i$\nobreakdash-colored vertices in $x$, then
 $n_i(x) < n_i(y)$ for the largest index $i$ with $n_i(x) \neq n_i(y)$.
\end{itemize}
\end{definition}
%

As remarked in~\cite{Sudholt2010b}, decreasing the number of vertices with the currently highest color (and not introducing yet a higher color) yields an improvement. If this number decreases to $0$, then the total number of colors has decreased.

We use the same local search operator as in~\cite{Sudholt2010b} called \emph{Grundy local search} (Algorithm~\ref{alg:Grundy-local-search}).
A vertex $v$ is called a \emph{Grundy vertex} if $v$ has the smallest color value not taken by any of its neighbors, formally
$
c(v) = \min\{i \in \{1, \dots, n\} \mid \forall w \in \neighborhood(v) \colon c(w) \neq i\}$,
where $\neighborhood(v)$ denotes the neighborhood of~$v$.
A coloring is called a \emph{Grundy coloring} if all vertices are Grundy vertices~\cite{Zaker2007}. 
Note that a Grundy coloring is always proper.
\begin{algorithm}[htb]
    \caption{Grundy local search~\cite{HedJacSri03}}
    \algsetup{indent=1.5em}
    \begin{algorithmic}[1]
        \WHILE{the current coloring is not a Grundy coloring}
        \STATE Choose a non-Grundy vertex $v$.
        \STATE Set $c(v) := \min\{i \in \{1, \dots, n\} \mid \forall w \in \neighborhood(v) \colon c(w) \neq i\}$.
        \ENDWHILE
    \end{algorithmic}
    \label{alg:Grundy-local-search}
\end{algorithm}

The analysis in~\cite{HedJacSri03} reveals that one step of the Grundy local search can only increase the color of a vertex if there is a conflict; otherwise the color of vertices can only decrease.
\cite{Sudholt2010b} point out that the application of Grundy local search can never worsen a coloring. If $y$ is the outcome of Grundy local search applied to~$x$ then $y \succeq x$.
If $x$ contains a non-Grundy node then $y$ is strictly better, i.e., $y \succeq x$ and $x \not\succeq y$.

We also introduce the \emph{Grundy number $\Gamma(G)$} of a graph~$G$ (also called \emph{first-fit chromatic number}~\cite{Balogh2008}) as the maximum number of colors used in any Grundy coloring. Every application of Grundy local search produces a proper coloring with color values at most~$\Gamma$.

We consider the \emph{Kempe chain} mutation operator defined in~\cite{Sudholt2010b}, which is based on so-called \emph{Kempe chain}~\cite{JenBja95} moves. This mutation exchanges two colors in a connected subgraph.
%
By $H_{ij}$ we denote the
set of all vertices colored~$i$ or~$j$ in~$G$. Then $H_j(v)$ is the
connected component of the subgraph induced by $H_{c(v)j}$ that contains~$v$.

\begin{algorithm}[htb]
    \caption{Kempe chain~\cite{Sudholt2010b}}
    \algsetup{indent=1.5em}
    \begin{algorithmic}[1]
        \STATE Choose $v \in V$ and $j \in \{1, \dots, \deg(v)+1\}$ uniformly at random.\!\!\!
        \STATE Let $i := c(v)$
        \FOR{all $u \in H_j(v)$}
            \STATE \textbf{if} $\col(u) = i$ \textbf{then} $\col(u) := j$ \textbf{else} $\col(u) := i$.
        \ENDFOR
    \end{algorithmic}
    \label{alg:Kempe-chain}
\end{algorithm}
The Kempe chain operator (Algorithm~\ref{alg:Kempe-chain}) is applied to a vertex $v$ and it exchanges the color of~$v$ (say $i$) with a color~$j$. We restrict the choice of~$j$ to the set $\{1, \dots, {\deg(v)+1}\}$ since larger colors will be replaced in the following Grundy local search. In the connected component $H_j(v)$ the colors $i$ and $j$ of all vertices are exchanged. As no conflict within $H_j(v)$ is created and $H_j(v)$ is not neighbored to any vertex colored $i$ or $j$, Kempe chains preserve feasibility.


An important point to note is that, when the current largest color is $c_{\max}$, Kempe chains are often most usefully applied to \emph{the neighborhood} of a $c_{\max}$-colored vertex~$v$. This can lead to a color in $v$'s neighborhood becoming a free color, and then the following Grundy local search will decrease the color of~$v$. In contrast, applying a Kempe chain to $v$ directly will spread color $c_{\max}$ to other parts of the graph, which might not be helpful.

\cite{Sudholt2010b} introduced a mutation operator called a \emph{color elimination} (Algorithm~\ref{alg:color-elimination}): it tries to eliminate a smaller color~$i$ in the neighborhood of a vertex~$v$ in one shot by trying to recolor all these vertices with another color~$j$ using parallel Kempe chains. Let $v_1, \dots, v_\ell$ be all $i$-colored neighbors of~$v$, for some number $\ell \ge 1$, then a Kempe chain move is applied to the union of the respective subgraphs, $H_j(v_1) \cup \dots \cup H_j(v_\ell)$.

\begin{algorithm}[hbt]
    \caption{Color elimination~\cite{Sudholt2010b}}
    \algsetup{indent=1.5em}
    \begin{algorithmic}[1]
        \STATE Choose $v \in V$ uniformly at random.
        \IF{$c(v) \ge 3$}
            \STATE Choose $i, j \in \{1, \dots, c(v)-1\}$, $i \neq j$, uniformly at random.
            \STATE Let $v_1, \dots, v_\ell$ enumerate all $i$\nobreakdash-colored neighbors of $v$.
            \FOR{all $u \in H_j(v_1) \cup \dots \cup H_j(v_\ell)$}
            \STATE \textbf{if} $\col(u) = i$ \textbf{then} $\col(u) := j$ \textbf{else} $\col(u) := i$.
        \ENDFOR
        \ENDIF
    \end{algorithmic}
    \label{alg:color-elimination}
\end{algorithm}

Iterated local search (ILS, Algorithm~\ref{alg:ILS}) repeatedly uses one of the aforementioned two mutations followed by Grundy local search. The mutation operator is not specified yet, but regarded as a black box. In the initialization every vertex $v$ receives an arbitrary color, which is then turned into a Grundy coloring by Grundy local search.

\renewcommand{\algorithmicloop}{\textbf{repeat forever}}
\begin{algorithm}[hbt]
    \caption{Iterated local search (\MA) ($x$)}
    \algsetup{indent=1.5em}
    \begin{algorithmic}[1]
        \STATE Replace $x$ by the result of Grundy local search applied to~$x$.
        \LOOP
            \STATE Let $y$ be the result of a \emph{mutation operator} applied to~$x$.
            \STATE Let $z$ be the outcome of Grundy Local Search applied to $y$.
            \STATE If $z \succeq x$ then $x := z$.
        \ENDLOOP
    \end{algorithmic}
    \label{alg:ILS}
\end{algorithm}

\subsection{Reoptimization Times}
\label{sec:dynamic-setting}

We consider the \emph{batch-update} model for dynamic graph coloring. That is, given a graph $G'=(V,E')$ and its proper coloring, we would like to find a proper coloring of $G=(V,E)$ which is obtained after a batch of up to\footnote{We say ``\emph{up to} $T$ edges'' instead of ``\emph{exactly $T$} edges'' as some negative results are easier to prove if just one edge is added.} $T$ edge insertions to $G'$.
We are interested in the reoptimization time, i.e., the number of iterations it takes to find a proper coloring of the current graph $G$, given a proper coloring of $G'$. How does the expected reoptimization time depend on $n$ and~$T$?
More precisely, we consider the \emph{worst case reoptimization time} to be the reoptimization time when considering the worst possible way of inserting up to $T$ edges into the graph.




Note that a bound for the reoptimization time can also yield a bound on the optimization time in the static setting for a graph $G=(V, E)$. This is because the static setting can be considered as a dynamic setting where we start with $n$ isolated vertices and then add all $T = |E|$ edges to the graph.

We point out that while we measure the number of iterations for all algorithms, the computational effort to execute one iteration may differ significantly between representations. RLS and \EA on bounded-size palettes only make small changes to the graph (in expectation).
For unbounded-size palettes, larger changes in the graph are possible. This presents a significant advantage for escaping from local optima and advancing towards the optimum, but it takes more computational effort.
The following theorem gives bounds on the computational complexity of executing one iteration of each algorithm in terms of elementary operations on a RAM machine.
\begin{theorem}
\label{the:execution-times-unaware}
Consider RLS, \EA and ILS on a connected graph $G=(V, E)$ with $|V| \ge 2$. Then
\begin{enumerate}
    \item one iteration of RLS can be executed in
    expected time $O(|E|/|V|)$,
    \item one iteration of the \EA can be executed in expected time $O(|E|/|V|)$, and
    \item one iteration of ILS with Kempe chains or color eliminations can be executed in time $O(|E|)$.
\end{enumerate}
\end{theorem}
In order to keep the paper streamlined, a proof for this theorem is given in Appendix \ref{app:some_proofs}. Note that, for graphs with $|E| = O(|V|)$, one iteration of RLS and the \EA can be executed in expected time $O(1)$, whereas one iteration of ILS can be executed in expected time $O(|V|) = O(n)$. For all connected graphs with at least two vertices, the upper bound for ILS is by a factor of $O(n)$ larger than the bounds for RLS and the \EA.

To be clear: Theorem~\ref{the:execution-times-unaware} is used to provide further context to the algorithms studied here. In the following theoretical results we will use the number of iterations as performance measure as customary in runtime analysis of randomized search heuristics and for consistency with previous work.

\subsection{Related work on (dynamic) graph coloring in the context of problem specific approaches}
\label{sec:related-work}

We remark that the coloring problems studied in this work are easy from a computational complexity point of view. A simple breadth-first search can be used to check in time $O(|V|+|E|)$ whether a graph $G = (V, E)$ is bipartite, i.e., 2-colorable, or not, and to find a proper 2-coloring if it is. Planar graphs can be colored with 4 colors (that is, even less than 5 colors) in time $O(|V|^2)$~\cite{Robertson1996}. The algorithm is quite complex and based on the proof of the famous Four Color Theorem.

For dynamically changing graphs, a number of dynamic graph coloring algorithms have been proposed. A general lower bound limits their efficiency: for any dynamic algorithm $\mathcal{A}$ that maintains a $c$-coloring of a graph, there exists a dynamic forest such that $\mathcal{A}$ must recolor at least $\Omega(|V|^{2/c(c-1)})$ vertices per update on average, for any $c\geq 2$ \cite{barba2017dynamic}.
For $c=2$, this gives a lower bound of $\Omega(|V|)$.
By a result in \cite{henzinger2020explicit} (which improves upon~\cite{SW18:coloring}), one can maintain an $O(\log n)$-coloring of a planar graph with \emph{amortized} polylogarithmic update time. There is a line of research on dynamically maintaining a $(\Delta+1)$-coloring of a graph with maximum degree at most $\Delta$~\cite{BCHN18:coloring,bhattacharya2019fully,henzinger2020constant} and the current best algorithm has $O(1)$ update time~\cite{bhattacharya2019fully,henzinger2020constant}.
In our setting of planar graphs with $\Delta \le 6$, this would only guarantee a proper coloring with 7 colors, though.

\section{Reoptimization Times on Bipartite Graphs}
\label{sec3}

We start off by considering bipartite graphs, i.e.\ 2-colorable graphs. For the bounded-size palette, we assume that only 2 colors are being used, i.e.\ $k=2$. We also consider unbounded-size palettes where the aim is to eliminate all colors larger than~2 from the graph.

\subsection{Paths and Binary Trees}

We first show that even adding a single edge can result in difficult symmetry problems. This can happen if two subgraphs are connected by a new edge, and then the coloring in one subgraph has to be inverted to find the optimum. Two examples for this are paths and binary trees.

\begin{theorem}
\label{theorem:path}
If adding up to $T$ edges completes an $n$-vertex path, the worst-case expected time for the \EA and RLS to rediscover a proper 2-coloring is $\Theta(n^3)$.
\end{theorem}
\begin{proof}
The claim essentially follows from the proofs of Theorems~3 and 5 in~\cite{Fischer2005} where the authors investigate an equivalent problem on cycle graphs. Hence, we just sketch the idea here. Imagine we link two properly colored paths of length $n/2$ each with an edge $(u,v)$ which introduces a single conflict.
The conflict splits the path into two paths that are properly colored and joined by a conflicting edge. Consider the length of the shortest properly colored path. As argued in~\cite{Fischer2005}, both RLS and \EA can either increase or decrease this length in fitness-neutral operations like recoloring one of the vertices involved in the conflict. If it has decreased to~1, the conflict has been propagated down to a leaf node where a single bit flip can get rid of it.
Fischer and Wegener calculate bounds for the expected number of steps until this number reaches its minimum 1. This is achieved by estimating the number of so-called relevant steps, which either increase or decrease the length of the shortest properly colored path. The probability for a relevant step is $\Theta(1/n)$. The expected number of relevant steps is $\Theta(n^2)$ since we have a fair random walk on states up to $n/2$. In summary, this results in a runtime bound of $\Theta(n^3)$.

Fischer and Wegener~\cite{Fischer2005} give an upper bound of $O(n^3)$ that holds for an arbitrary initialization, hence the upper bound holds for arbitrary values of~$T$.
\end{proof}


\newcommand{\treenode}[3]{node[#1, label=right:#3]{#2}}
        \tikzset{label distance=-5pt}
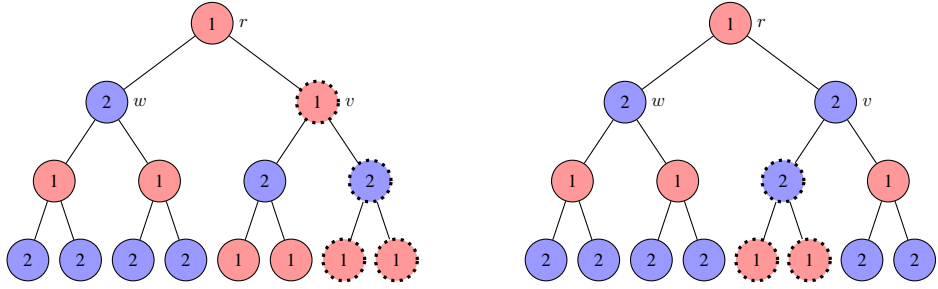
\begin{figure}[tb]
    \centering
        \tikzstyle{level 1}=[sibling distance=4cm]
        \tikzstyle{level 2}=[sibling distance=2.0cm]
        \tikzstyle{level 3}=[sibling distance=1.0cm]
        \tikzstyle{level 4}=[sibling distance=0.5cm]
    \begin{tikzpicture}[nodes={draw,circle,fill=gray!20}, every node/.append style={minimum width=10, inner sep=5},scale=0.7, transform shape]
        \tikzstyle{vertex}=[draw,circle,inner sep=0cm,minimum size=.4cm];
        \tikzstyle{edge}=[black];
        \tikzstyle{toflip}=[very thick, dotted];

    \node[color1,label=right:$r$]{1}
        child{\treenode{color2}{2}{$w$}
            child{\treenode{color1}{1}{}
                child{\treenode{color2}{2}{}}
                child{\treenode{color2}{2}{}}
            }
            child{\treenode{color1}{1}{}
                child{\treenode{color2}{2}{}}
                child{\treenode{color2}{2}{}}
            }
        }
        child{\treenode{color1,toflip}{1}{$v$}
            child{\treenode{color2}{2}{}
                child{\treenode{color1}{1}{}}
                child{\treenode{color1}{1}{}}
            }
            child{\treenode{color2,toflip}{2}{}
                child{\treenode{color1,toflip}{1}{}}
                child{\treenode{color1,toflip}{1}{}}
            }
        }
    ;

    \begin{scope}[xshift=280]
    \node[color1,label=right:$r$]{1}
        child{\treenode{color2}{2}{$w$}
            child{\treenode{color1}{1}{}
                child{\treenode{color2}{2}{}}
                child{\treenode{color2}{2}{}}
            }
            child{\treenode{color1}{1}{}
                child{\treenode{color2}{2}{}}
                child{\treenode{color2}{2}{}}
            }
        }
        child{\treenode{color2}{2}{$v$}
            child{\treenode{color2,toflip}{2}{}
                child{\treenode{color1,toflip}{1}{}}
                child{\treenode{color1,toflip}{1}{}}
            }
            child{\treenode{color1}{1}{}
                child{\treenode{color2}{2}{}}
                child{\treenode{color2}{2}{}}
            }
        }
    ;
    \end{scope}

    \end{tikzpicture}
\caption{A complete binary tree with a worst-case coloring in $A_0$ (left) and a coloring in $A_1$ (right). Flipping the dotted vertices is sufficient to make a transition from $A_0$ to $A_1$ and from $A_1$ to a proper coloring, respectively.}
    \label{fig:binary-trees}
\end{figure}

\begin{theorem}
\label{theorem:tree}
If adding an edge completes an $n$-vertex complete binary tree, the worst-case expected time for the \EA to rediscover a proper 2-coloring is $\Theta\big(n^{(n+1)/4}\big)$. For both static and dynamic settings, RLS is unable to find a proper 2-coloring in the worst case.
\end{theorem}
\begin{proof}
The proof uses and refines arguments from~\cite{Sudholt2005}.
Let $e = \{r,v\}$ be the added edge with $r$ being the root of the $n$-vertex complete binary tree. If $c(r) \neq c(v)$ we are done and the coloring is already a proper 2-coloring. Hence, we assume that $c(r) = c(v)$ and there is exactly one conflict. This situation is a worst-case situation in vertex-coloring of complete binary trees, since many vertices must be recolored in the same mutation to produce an accepted candidate solution (see Figure~\ref{fig:binary-trees} (left)).
Let $\mathrm{OPT}$ be the set of the two possible proper colorings and let $A_i$, for $0 \le i \le \log(n)-1$ be the set of colorings with one conflict such that the parent vertex of the conflicting edge has (graph) distance~$i$ to the root. We have  $\sum_{i=0}^{\log(n)-1} |A_i| = 2n-2$ since we can choose the position of the conflicting edge among $n-1$ edges and there are two possible colors for its vertices.
By the same argument, $|A_0|=4$ and $|A_1|=8$.

Starting with a coloring $x \in A_0$, the probability of reaching $\mathrm{OPT}$ in one mutation is at most $n^{-(n-1)/2} + n^{-(n+1)/2} = O(n^{-(n-1)/2})$ since all vertices on either side of the conflicting edge must be recolored in one mutation.
The probability of reaching $A_1$ in one mutation is $\Omega(n^{-(n+1)/4})$ since a sufficient condition is to flip $v$ and all the vertices in one of $v$'s subtrees (see Figure~\ref{fig:binary-trees} (left)). Since each subtree of $v$ has $(n-3)/4$ vertices, this means flipping $1 + (n-3)/4 = (n+1)/4$ many vertices. This probability is also $O(n^{-(n+1)/4})$ since the only other way to create some coloring in~$A_1$ is to flip $r$, the sibling of~$v$ (that we call~$w$), and one of $w$'s subtrees.
The probability to reach any solution in
$\bigcup_{i=2}^{\log(n)-1} A_i$ is $O(n^{-(n+1)/4})$ as well since more than $(n+1)/4$ vertices would have to flip and there are $2n - 2 - |A_0| - |A_1|=2n-14$ solutions in $\bigcup_{i=2}^{\log(n)-1} A_i$. This implies the claimed lower bound as the probability to escape from $A_0$ in one mutation is
$\Theta(n^{-(n+1)/4})$.

To show the claimed upper bound, we argue that in $\Theta(n^{(n+1)/4})$ expected time we do escape from $A_0$. If $\mathrm{OPT}$ is reached, we are done. Hence, we assume that $\bigcup_{i=1}^{\log(n)-1} A_i$ is reached. For each coloring in this set, there is a proper coloring within Hamming distance at most $(n-3)/4$ since, if $\{u, v\}$ denotes the conflicting edge with $u$ being the parent of~$v$, it is sufficient to recolor the subtree at~$v$ and this subtree has at most $(n-3)/4$ vertices (see Figure~\ref{fig:binary-trees} (right)).
Thus, the expected time to either reach $\mathrm{OPT}$ or to go back to $A_0$ is $O(n^{(n-3)/4})$. Since at least $(n+1)/4$ vertices would have to flip to go back to $A_0$ (and $|A_0|=O(1)$), the conditional probability to go back to $A_0$ is at most $O(1/n)$. If this happens, we repeat the above arguments; this clearly does not change the asymptotic runtime and we have shown an upper bound of $O(n^{(n+1)/4})$.

It is obvious from the above that RLS is unable to leave $A_0$ and hence it fails in both static and dynamic settings when starting with a worst-case initialization.
\end{proof}


In the above two examples, the reoptimization time is at least as large as the optimization time from scratch.
In fact, our dynamic setting even allows us to create a worst-case initial coloring that might not typically occur with random initialization. Theorem~\ref{theorem:path} gives a rigorous lower bound of order~$n^3$ as after adding an edge connecting two paths of $n/2$ vertices each, we start the last ``fitness level'' with a worst-case initial setup.  \cite{Fischer2005} were only able to show a lower bound under additional assumptions. Also in~\cite{Sudholt2005} the probability of reaching the worst-case situation described in Theorem~\ref{theorem:tree} was very crudely bounded from below by $\Omega(2^{-n})$. Our lower bounds for dynamic settings are hence a bit tighter and/or more rigorous than those for the static setting.

The reason for the large reoptimization times in the above cases is because for the \EA and \RLS mutations occur locally, and they struggle in solving symmetry problems where large parts of the graph need to be recolored.
Mutation operators in ILS like Kempe chains and color eliminations operate more globally, and can easily deal with the above settings.
\begin{theorem}
\label{the:ils-on-paths}
Consider a dynamic graph that is a path after a batch of up to $T$ edge insertions.
The expected time for ILS with Kempe chains to rediscover a proper 2-coloring on paths is $O(n)$.
\end{theorem}
\begin{proof}
The statement about paths follows from~\cite[Theorem~1]{Sudholt2010b} as the expected time to 2-color a path is $O(n)$ in the static setting. (It is easy to see that the proof holds for arbitrary initial colorings.)
\end{proof}

\begin{theorem}
\label{the:ils-on-trees}
Consider a dynamic graph that is a binary tree after a batch of up to $T$ edge insertions.
The expected time for ILS with either Kempe chains or color eliminations to rediscover a proper 2-coloring or to find a proper 2-coloring in the static setting (where $T=n-1$) is $O(n \log^+ T)$.
\end{theorem}

The upper bound of $O(n \log^+ T)$ is an improvement over the bound $O(n \log n)$ from the conference version of this paper~\cite[Theorem~3.3]{BNPS2019}.
The proof uses the following structural insights that apply to all graphs and will also prove useful in the analysis of planar graphs in Section~\ref{sec4}. By the design of the selection operator, the number of $(\Delta+1)$-colored vertices is non-increasing over time.
We shall show that also the number of vertices colored $\Delta$ or $\Delta+1$ is non-increasing.

The following lemma shows that a Kempe chain operation or color elimination can only increase the number of $\Delta$-colored vertices by at most~1.

\begin{lemma}
\label{lem:colors-in-H-cDelta}
Consider a Grundy-colored graph with maximum degree~$\Delta$. Then every Kempe chain operation and every color elimination can only increase the number of $\Delta$-colored vertices by at most~1.
\end{lemma}
\begin{proof}
We first consider Kempe chains and distinguish between two cases: the Kempe chain involves colors $\Delta$ and $\Delta+1$ and the case that it involves colors $\Delta$ and a smaller color $c < \Delta$. Kempe chains involving two colors other than $\Delta$ cannot change the number of $\Delta$-colored vertices (albeit this may still happen in a subsequent Grundy local search, if $(\Delta+1)$-colored vertices are recolored with color~$\Delta$). We start with the case of colors $\Delta$ and $\Delta+1$.

Assume that there exists a vertex $v$ that is being recolored from $\Delta+1$ to~$\Delta$ in the Kempe chain (if no such vertex exists, the claim holds trivially). Since the coloring is a Grundy coloring, $v$ must have vertices of all colors in $\{1, \dots, \Delta\}$ in its neighborhood, and there can only be one vertex of each color (owing to the degree bound~$\Delta$). Let $w$ be the $\Delta$-colored vertex and note that $w$ must have all colors from 1 to $\Delta-1$ in its neighborhood. Thus, $w$ must have neighbors $w_1, \dots, w_{\Delta-1}$ such that $w_i$ is colored~$i$. Since $w$ also has $v$ as its neighbor, $w$ cannot have any further neighbors apart from $v, w_1, \dots, w_{\Delta-1}$; in particularly, $w$ cannot have any further $(\Delta+1)$-colored vertices as neighbors. Hence the subgraph $H_{\Delta (\Delta+1)}$ induced by vertices colored~$\Delta$ or $\Delta+1$ contains $\{v, w\}$ as a connected component and a Kempe chain on this component will simply swap the colors of $v$ and $w$ without increasing the number of $\Delta$-colored vertices.

Now assume that the other color is $c < \Delta$. We show that in the subgraph
$H_{c\Delta}$ induced by vertices colored~$c$ or $\Delta$, the
number of $c$-colored vertices in every connected component of
$H_{c\Delta}$ is at most 1 larger than the number of
$\Delta$-colored vertices. This implies the claim since a Kempe chain operation swaps colors $c$ and $\Delta$ in one connected component of $H_{c\Delta}$.

Every $\Delta$-colored vertex~$w$ needs to have colors $\{1,
\dots, \Delta-1\}$ in its neighborhood since the coloring is a Grundy
coloring. Since the maximum degree is~$\Delta$, $w$ can have at
most two $c$-colored neighbors.

Consider a connected component of $H_{c\Delta}$ that contains a
$c$-colored vertex~$v$ (if no such vertex exists, the claim holds trivially). Imagine the breadth-first search (BFS) tree
generated by running BFS in $H_{c\Delta}$ starting at~$c$. Note
that colors are alternating at different depths of the BFS tree, with
$\Delta$-colored vertices at odd depths and $c$-colored vertices at
even depths from the root. For odd depths $d$, all $\Delta$-colored vertices can
only have 1 $c$-colored vertex at depth~$d+1$ since they are
already connected to one $c$-colored vertex at depth~$d-1$. Hence
there are at least as many $\Delta$\nobreakdash-colored vertices at depth~$d$
as $c$-colored vertices at depth~$d+1$. Using this argument for all
odd values of~$d$ and noting that the root vertex~$v$ is
$c$-colored proves the claim.

For color eliminations, recall that the parameters are two colors that are smaller than the color of the selected vertex. So a color~$\Delta$ can only be involved if the selected vertex~$v$ has color~$\Delta+1$. Since every color value $\{1, \dots, \Delta\}$ appears exactly once in the neighborhood of~$v$, a color elimination with parameters $i, j$ boils down to one Kempe chain with colors $i$ and $j$. Then the claim follows from the statement on Kempe chains.
\end{proof}


Lemma~\ref{lem:colors-in-H-cDelta} implies that the number of vertices colored $\Delta$ or $\Delta+1$ can never increase in an iteration of ILS.
\begin{lemma}
\label{lem:number-of-Delta-and-Delta-plus-1-non-increasing}
On every Grundy-colored graph, an iteration of ILS with either Kempe chains or color eliminations does not increase the number of vertices colored either $\Delta$ or $\Delta+1$.
\end{lemma}
\begin{proof}
The number of $(\Delta+1)$-colored vertices is non-increasing by design of the selection operator. Moreover, the number of $\Delta$-colored vertices can only increase if the number of $(\Delta+1)$-colored vertices decreases at the same time. If there are no $(\Delta+1)$-colored vertices, the number of $\Delta$-colored vertices is non-increasing.
Hence we only need to consider the case where there is at least one $(\Delta+1)$-colored vertex.

The proof of Lemma~\ref{lem:colors-in-H-cDelta} revealed that every Kempe chain or color elimination can only increase the number of $\Delta$-colored vertices by~1. Moreover, this can only happen if a Kempe chain affects a connected component $C$ of $H_{c\Delta}$, for a color $c < \Delta$, such that $C$ has one more $c$-colored vertex than $\Delta$-colored vertices. For this operation to be accepted by selection, the following Grundy local search must reduce the number of $(\Delta+1)$-colored vertices by at least~1. 


Consider one $(\Delta+1)$-colored vertex~$v$ whose color decreases. If the new color is smaller than~$\Delta$, $v$ does not increase the number of $\Delta$-colored vertices. If its new color is~$\Delta$, then we claim that there must exist another vertex $w \notin C$ whose color decreases from $\Delta$ to a smaller color.
Note that $v$ can only be recolored~$\Delta$ if $\Delta$ becomes a free color for~$v$, that is, the unique $\Delta$-colored neighbor~$w$ of~$v$ is being recolored (recall that all colors $\{1, \dots, \Delta\}$ appear once in the neighborhood of~$v$). The proof of Lemma~\ref{lem:colors-in-H-cDelta} showed that $w \notin C$ as otherwise $w$ would have more than $\Delta$ neighbors. It also showed that $v$ and $w$ cannot have any edges to other vertices colored $\Delta$ or $\Delta+1$. Hence if there are $\ell > 1$ vertices $v_1, \dots, v_\ell$ whose color decreases from $\Delta+1$ to~$\Delta$ then there are $\ell$ vertices $w_1, \dots, w_\ell \in G \setminus C$ whose color decreases from $\Delta$ to a smaller color. This implies the claim.
%
\end{proof}

When adding edges in the unbounded-size palette setting, Grundy local search will repair any conflicts introduced in this way by increasing colors of vertices incident to conflicts.
The following lemma states that the number of colors being increased is bounded by the number of inserted edges.
\begin{lemma}
\label{lem:number-of-increased-colors}
When inserting at most $T$ edges into a graph that is Grundy colored, the following Grundy local search will only recolor up to $T$ vertices.
\end{lemma}
\begin{proof}
As shown in \cite[Lemma~3]{HedJacSri03}, one step of the Grundy local search can only increase the color of a vertex if it is involved in a conflict. Otherwise, the color of vertices can only decrease. If a vertex $v$ is involved in a conflict and subsequently assigned the smallest free color, all conflicts at $v$ are resolved and $v$ will never be touched again during Grundy local search~\cite[Lemma~4]{HedJacSri03} since further steps of the Grundy local search cannot create new conflicts. Hence after at most $T$ steps, Grundy local search stops with a Grundy coloring.
\end{proof}

{With the above lemmas, we are ready to prove Theorem \ref{the:ils-on-trees}.}

\begin{proof}[Proof of Theorem~\ref{the:ils-on-trees}]
The Grundy number of binary trees is at most $\Gamma \le {\Delta +1} \le 4$. By Lemma~\ref{lem:number-of-increased-colors}, the number of vertices colored 3 or~4 is at most~$T$.

By design of our selection operator, the number of 4-colored vertices is non-increasing over time. For every 4-colored vertex~$v$ there must be a Kempe chain operation recoloring a neighboring vertex whose color only appears once in the neighborhood of~$v$. If there are $i$ 4-colored vertices, the probability of reducing this number is $\Omega(i/n)$ and the expected time for color~4 to disappear is $O(n) \cdot \sum_{i=1}^T 1/i = O(n \log^+ T)$.

Since the number of vertices colored 3 or 4 cannot increase by Lemma~\ref{lem:number-of-Delta-and-Delta-plus-1-non-increasing}, once all 4-colored vertices are eliminated, there will be at most $T$ 3-colored vertices, and the time to eliminate these is bounded by $O(n \log^+ T)$ by the same arguments as above.
\end{proof}

\subsection{A Bound for General Bipartite Graphs}

\cite{Sudholt2010b} showed that ILS with color eliminations can color every bipartite graph efficiently, in expected $O(n^2 \log n)$ iterations~\cite[Theorem~3]{Sudholt2010b}. The main idea behind this analysis was to show that the algorithm can eliminate the highest color from the graph by applying color eliminations to all such vertices. The expected time to eliminate the highest color is $O(n \log n)$, and we only have to eliminate at most $O(n)$ colors. In fact, the last argument can be improved by considering that in every Grundy coloring of a graph~$G$ the largest color is at most $\Gamma(G)$. This yields an upper bound of $O(\Gamma(G)n \log n)$ for both static and dynamic settings.

The following result gives an additional bound of $O(\sqrt{T}n \log n)$, showing that the number $T$ of added edges can have a sublinear impact on the expected reoptimization time.
\begin{theorem}
\label{the:ils-ce-rediscover-time}
Consider a dynamic graph that is bipartite after a batch of up to $T$ edge insertions. Let $\Gamma$ be the Grundy number of the resulting graph. Then ILS with color eliminations re-discovers a proper 2-coloring in expected
$O(\min\{\sqrt{T}, \Gamma\} n \log n)$ iterations.
\end{theorem}
\begin{proof}
Consider the connected components of the original graph. If an edge is added that runs within one connected component, it cannot create a conflict. This is because the connected component is properly 2-colored, with all vertices of the same color belonging to the same set of the bipartition. Since the graph is bipartite after edge insertions, the new edge must connect two vertices of different colors.
Hence added edges can only create a conflict if they connect two different connected components that are colored inversely to each other.

Consider the subgraph induced by the added edges that are conflicting, and pick a connected component~$C$ in this subgraph. Note that all vertices in~$C$ have the same color $c \in \{1, 2\}$ before Grundy local search is applied. Now Grundy local search will fix these conflicts by increasing the colors of vertices in~$C$. We bound the value of the largest color $c_{\max}$ used and first consider the case where the largest color is $c_{\max} \ge 4$.
For Grundy local search to assign a color~$c_{\max} \ge 4$ to a vertex~$v \in C$, all colors $1, \dots, c_{\max}-1$ must occur in the neighborhood of~$v$ in the new graph. In particular, $C$ must contain vertices $v_3, v_4, \dots, v_{c_{\max}-1}$ respectively colored $3, \dots, c_{\max}-1$ that are neighbored to~$v$. This implies that $c_{\max}-3$ edges incident to~$v$, connecting~$v$ to a smaller color, must have been added during the dynamic change. Applying the same argument to $v_3, v_4, \dots, v_{c_{\max}-1}$ yields that there must be at least $\sum_{j=1}^{c_{\max}-3} j = (c_{\max}-3)(c_{\max}-2)/2$ inserted edges in~$C$. Thus $(c_{\max}-3)(c_{\max}-2)/2 \le T$, which implies $(c_{\max}-3)^2 \le 2T$ and this is equivalent to $c_{\max} \le \sqrt{2T} + 3$.
Also $c_{\max} \le \Gamma$ by definition of the Grundy number.

Now we can argue as in~\cite[Theorem~3]{Sudholt2010b}: the largest color can be eliminated from any bipartite graph in expected time $O(n \log n)$. (Note that these color eliminations can increase the number of vertices colored with large colors, so long as the number of the vertices with the largest color decreases.) Since at most $c_{\max}-2$ colors have to be eliminated, a bound of $O(c_{\max} n \log n)$ follows. Plugging in $c_{\max} = O(\min\{\sqrt{T}, \Gamma\})$ completes the proof.

If Grundy local search uses a largest color of $c_{\max} \le 3$ an $O(n \log n)$ bound follows as for $c_{\max}=3$ only one color has to be eliminated and $c_{\max} \le 2$ implies that a proper coloring has already been found.
%
%
\end{proof}

For graphs with Grundy number $\Gamma \le 4$, which includes binary trees, star graphs, paths and cycles, the bound improves to $O(n \log^+ T)$.
\begin{theorem}
\label{the:general-n-log-T-bound-for-ils-with-ce}
Consider a dynamic graph that is bipartite after a batch of up to $T$ edge insertions.
If no end point of an added edge is neighbored to an end point of another added edge, or if $\Gamma \le 4$, the expected time to re-discover a proper 2-coloring is $O(n \log^+ T)$.
If only one conflicting edge is added, the expected time is~$\Theta(n)$.
%
\end{theorem}
\begin{proof}
If no end point of an added edge is neighbored to an end point of another added edge, Grundy local search will only create colors up to~3.
This is because Grundy local search will only increase the color of end points of added edges, and the condition implies that the colors of neighbors of all end points will remain fixed. Hence, Grundy local search will recolor vertices independently from each other. If $\Gamma \le 4$, the largest color value is~4.
Lemma~\ref{lem:number-of-Delta-and-Delta-plus-1-non-increasing} states that the number of vertices colored 3 or 4 cannot increase.

Following~\cite[Theorem~3]{Sudholt2010b}, while there are $i$ vertices colored~4, a color elimination choosing such a vertex will lead to a smaller free color, reducing the number of 4-colored vertices. The expected time for this to happen is at most $n/i$, hence the total expected time to eliminate all color-4 vertices is at most $\sum_{i=1}^T n/i = O(n \log^+ T)$. The same argument then applies to all 3-colored vertices.

If only one conflicting edge is inserted ($T=1$) then there will be one 3\nobreakdash-colored vertex~$v$ after Grundy local search, and a proper 2-coloring is obtained by applying a color elimination to~$v$. The expected waiting time for choosing vertex~$v$ is $\Theta(n)$.
\end{proof}

\label{subsec:infeasible_k2}
\begin{figure}[tb]
    \centering
    \begin{tikzpicture}[xscale=0.5,yscale=0.45]
        \tikzstyle{vertex}=[draw,circle,inner sep=0cm,minimum size=.4cm];
        \tikzstyle{edge}=[black];

        \draw (0, -3.5) node [vertex,color2] (center) {2};

        \foreach \i in {1,...,5} {
            \node [vertex,color1] (secondb\i) at (3, -\i) {1};
            \path (center) edge [edge] (secondb\i);
            \node [vertex,color2] (firstb\i) at (6, -\i) {2};
            \path (secondb\i) edge [edge] (firstb\i);
        }
            \node [vertex,color2] (seconda6) at (3, -6) {2};
            \path (center) edge [edge, dashed] (seconda6);
            \node [vertex,color1] (firsta6) at (6, -6) {1};
            \path (seconda6) edge [edge] (firsta6);
    \end{tikzpicture}
    \qquad
    \begin{tikzpicture}[xscale=0.5,yscale=0.45]
        \tikzstyle{vertex}=[draw,circle,inner sep=0cm,minimum size=.4cm];
        \tikzstyle{edge}=[black];

        \draw (0, -3.5) node [vertex,color3] (center) {3};

        \foreach \i in {1,...,5} {
            \node [vertex,color1] (secondb\i) at (3, -\i) {1};
            \path (center) edge [edge] (secondb\i);
            \node [vertex,color2] (firstb\i) at (6, -\i) {2};
            \path (secondb\i) edge [edge] (firstb\i);
        }
            \node [vertex,color2] (seconda6) at (3, -6) {2};
            \path (center) edge [edge, dashed] (seconda6);
            \node [vertex,color1] (firsta6) at (6, -6) {1};
            \path (seconda6) edge [edge] (firsta6);
    \end{tikzpicture}
    \caption{Depth-2 star with $n=13$ vertices. The dashed line indicates the added edge. Left: coloring with a bounded-size palette, right: coloring after Grundy local search with an unbounded-size palette.}
    \label{fig:worst-case-for-Kempe-chains}
\end{figure}
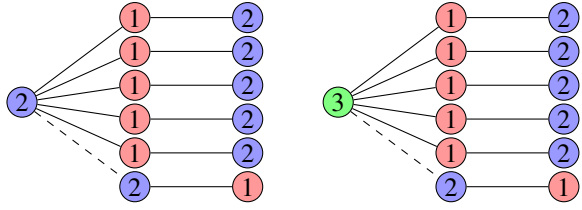
\vspace*{-0.3cm}

\subsection{A Worst-Case Graph for Kempe Chains}

While ILS with color eliminations efficiently reoptimizes all bipartite graphs, for ILS with Kempe chains there are bipartite graphs where even adding a single edge connecting a tree with an isolated edge can lead to exponential times.
\begin{theorem}
\label{the:worst-case-for-Kempe-chains}
For every $n \equiv 1 \bmod 4$ there is a forest $T_n$ with $n$ vertices such that for every feasible 2-coloring \MA with Kempe chains needs $\Theta(2^{n/2})$ generations in expectation to re-discover a feasible 2-coloring after adding an edge.
\end{theorem}
\begin{proof}
Choose $T_n$ as the union of an isolated edge $\{u, v\}$ where $c(u) = 2$ and $c(v) = 1$ and a tree where the root $r$ has $N-1 := {(n-3)/2}$ children and every child has exactly one leaf (cf.\ Figure~\ref{fig:worst-case-for-Kempe-chains}). This graph was also used in~\cite{Sudholt2010b} as an example where ILS with Kempe chains fail in a static setting. Since $n \equiv 1 \bmod 4$, $N$ is an even number.
Every feasible 2-coloring will color the root and the leaves in the same color and the root's children in the remaining color. Assume the root and leaves are colored 2 as the other case is symmetric.
Now add an edge $\{r, u\}$ to the graph. This creates a star of depth 2 (termed the \emph{depth-2 star} in the following) where the root is the center and the root now has $N$ children.

This creates a conflict at $\{r, u\}$ that is being resolved by recoloring one of these vertices to color~3 in the next Grundy local search. With probability $1/2$, this is the root~$r$.

From this situation, any Kempe chain affecting any vertex in $V \setminus \{r\}$ can swap the colors on an edge incident to a leaf.
Let $X_0, X_1, \dots$ denote the random number of leaves colored~1, starting with $X_0 = 1$. We only consider steps in which this number is changed; note that the probability of such a change is $\Theta(1)$ as every Kempe chain on any vertex except for the root changes $X_t$ if an appropriate color value is chosen.
There are $N := (n-1)/2$ leaves and the number of 1-colored leaves performs a random walk biased towards $N/2$: $\Prob{X_{t+1} = X_t+1 \mid X_t} = (N-X_t)/N$ and $\Prob{X_{t+1} = X_t-1 \mid X_t} = X_t/N$. This process is known as the Ehrenfest urn model: imagine two urns labelled 1 and 2 that together contain $N$ balls. In each step, we pick a ball uniformly at random and move it to the other urn. If $X_t$ denotes the number of balls in urn~1 at time~$t$, we obtain the above transition probabilities.\footnote{This simple model was originally proposed to describe the process of substance exchange between two bordering containers of equal size which are separated by a permeable membrane. Consider $N$ particles spread across the containers and denote by $X_t$ the number of particles in the left container \wlo at time $t$. In each step one particle is chosen uniformly at random and swaps sides.}

When $X_t \in \{0, N\}$ then a proper 2-coloring has been found. As long as $X_t \in \{2, \dots, N-2\}$, all Kempe chain moves involving the root will be rejected as the number of 3-colored vertices would increase.
While $X_t \in \{1, N-1\}$ a Kempe chain move recoloring the root with the minority color will be accepted. This has probability $1/n \cdot 1/(N-1) = \Theta(1/N^2)$ (as the color is chosen uniformly from $\{1, \dots, \deg(r)+1\}$) and then the following Grundy local search will produce a proper 2-coloring. Also considering possible transitions to neighbouring states $0$ or~$N$, while $X_t \in \{1, N-1\}$ the conditional probability that a proper 2-coloring is found before moving to a state $X_t \in \{2, N-2\}$ is $\Theta(1/N)$.

For the Ehrenfest model it is known that the expected time to return to an initial state of $1$  is
$1!(N-1)!/N! \cdot 2^N = 2^N/N$~\cite[equation (66)]{Kac1947}. It is easy to show that this time remains in $\Theta(2^N/N)$ when considering $N-1$ as a symmetric target state, and when conditioning on traversing states $\{2, \dots, N-2\}$. A rigorous proof for this statement is given in the Lemma~\ref{lem:Ehrenfest} stated after this proof.

Along with the above arguments, this means that such a return in expectation happens $\Theta(N)$ times before a proper 2-coloring is found. This yields a total expectation of $\Theta(2^N) = \Theta(2^{n/2})$.
\end{proof}

\begin{lemma}
\label{lem:Ehrenfest}
Consider the Ehrenfest urn model with $N$ balls spread across two urns 1 and 2, in which at each step a ball is picked uniformly at random and moved to the other urn.
Describing the current state as the number of balls in urn~1, when starting in either state~1 or state~$N-1$, the expected time to return to a state from $\{1, N-1\}$ via states in $\{2, \dots, N-2\}$ is $\Theta(2^N/N)$.
\end{lemma}
\begin{proof}
Let $T_{a \to b}$ denote the first-passage time from a state~$a$ to a state~$b$ and $T_{a \to B}$ for a set $B$ denote the first-passage time from $a$ to any state in~$B$.
For the Ehrenfest model it is known that the expected time to return to a state of $1$ from a state of~$1$ is
$\E{T_{1 \to 1}} = 1!(N-1)!/N! \cdot 2^N = 2^N/N$~\cite[equation (66)]{Kac1947}.
The Ehrenfest model starting in state~1 can return to state~1 either via state 0 or larger states. Since the former takes exactly 2 steps, the expected return time via larger states is also $\Theta(2^N/N)$ by the law of total expectation.

From state~$N/2$, by symmetry, there are equal probabilities of reaching state~1 or state~$N-1$ when we first reach a state from $\{1, N-1\}$. If state $N-1$ is reached, the model needs to return to $N/2$ and move from $N/2$ to 1 in order to reach state~$1$. This leads to the recurrence
\begin{align*}
T_{N/2 \to 1} =\;& T_{N/2 \to \{1, N-1\}} + \frac{1}{2} \cdot (T_{N-1 \to N/2} + T_{N/2 \to 1})\\
=\;& T_{N/2 \to \{1, N-1\}} + \frac{1}{2} \cdot (T_{1 \to N/2} + T_{N/2 \to 1})
\end{align*}
which is equivalent to
\[
 \frac{1}{2} \cdot T_{1 \to N/2} + T_{N/2 \to \{1, N-1\}} = \frac{1}{2} \cdot T_{N/2 \to 1} .
\]
Let $A$ be the event that the model, when starting in state~1, passes through state~$N/2$ before reaching a state from $\{1, N-1\}$ again. Then
\begin{align*}
& \E{T_{1 \to \{1, N-1\}}}\\
=\;& \Prob{A} \cdot \E{T_{1 \to \{1, N-1\}} \mid A} + \Prob{\overline{A}} \cdot \E{T_{1 \to \{1, N-1\}} \mid \overline{A}}\\
 =\;&\Prob{A} \cdot \E{T_{1 \to N/2} + T_{N/2 \to \{1, N-1\}}} + \Prob{\overline{A}} \cdot \E{T_{1 \to 1} \mid \overline{A}}\\
 =\;&\Prob{A} \cdot \E{\frac{T_{1 \to N/2} + T_{N/2 \to 1}}{2}} + \Prob{\overline{A}} \cdot \E{T_{1 \to 1} \mid \overline{A}}\\
 =\;&\Theta(\E{T_{1 \to 1}}) = \Theta(2^N/N).\qedhere
\end{align*}
\end{proof}

It is interesting to note that the worst-case instance for Kempe chains is easy for all other considered algorithms.

\begin{theorem}
\label{the:color-eliminations-on-worst-case-tree}
On a graph where adding up to $T$ edges completes a depth\nobreakdash-2 star, ILS with color eliminations rediscovers a proper 2-coloring in expected time $O(n \log^+ T)$. \end{theorem}
\begin{proof}
We argue that the graph's Grundy number is $\Gamma = 3$ as then the claim follows from Theorem~\ref{the:general-n-log-T-bound-for-ils-with-ce}.
Since all vertices but the root have degree at most~2, their colors must be at most~3. Assume for a contradiction that the root has a color larger than~3. Then there must be a child~$v$ of color~3. But then $v$ has a free color in $\{1, 2\}$, contradicting a Grundy coloring. Hence also the root must have color at most~$3$, completing the proof that $\Gamma=3$.
\end{proof}

\begin{theorem}
\label{the:rls-on-worst-case-tree-static}
On the depth-2 star RLS and \EA both have expected optimization time $O(n \log n)$ in the static setting and $O(n \log^+ T)$ to rediscover a proper 2-coloring after adding up to $T$ edges.
\end{theorem}

\begin{proof}
First note that any conflict can be resolved by one or two mutations. The latter is necessary in the unfavourable situation of $\{r,u\}, \{u,v\} \in E$, $r$ being the root, with $c(r) = 2 = c(u)$ and $c(v) = 1$. Then both $u$ and $v$ need to be recolored simultaneously or in sequence. We show that every conflict has a constant probability of being resolved within the next $n$ steps. Let $X_t$ denote the number of conflicts at time $t \in \mathbb{N}_0$. If $X_t > 0$, the probability of improvement within $n$ steps is at least
\[
    p \geq \frac{1}{2} \cdot \binom{n}{2} \cdot \left(\frac{1}{n}\right)^2 \cdot \left(\left(1 - \frac{1}{n}\right)^{n-1}\right)^2 \cdot \left(1 - \frac{2}{n}\right)^{n-2} \\
    \geq \frac{(n-1)}{4ne^4} = \Omega(1).
\]
Here, the term $1/2 \cdot \binom{n}{2}$ describes all combinations of two relevant mutations concerning nodes $u$ and $v$ in sequence. The next two factors indicate that in the selected steps both $u$ and $v$ are recolored and all remaining nodes are left apart. Finally, the last factor is the probability of not mutating both vertices in the remaining $n-2$ steps.
Note that for \RLS the penultimate factor disappears. Hence, the expected number of conflicts after $n$ steps is
\begin{align*}
    E(X_{t + n} \, | \, X_t)
    & \leq X_t - X_t p
    \leq X_t - X_t \cdot \frac{(n-1)}{4ne^4}
    = X_t \cdot \left(1 - \frac{(n-1)}{4ne^4}\right)
\end{align*}
and we obtain an expected multiplicative drift of
\[
    E(X_t - X_{t+n}\,|\, X_t) \geq X_t - X_t \cdot \left(1 - \frac{(n-1)}{4ne^4}\right) = X_t \frac{(n-1)}{4ne^4}.
\]
Applying the multiplicative drift theorem~\cite{Doerr2012} yields an upper bound of
$
    \frac{8e^2}{1 + 1/n} \log(1 + x_{\max}) = O(\log^{+} x_{\max})
$
for the expected number of phases. Here, $x_{\max} \le n$ in the static setting and $x_{\max} \le T$ in the dynamic setting denotes the maximum number of conflicts. Hence, the runtime bounds are $O(n \log n)$ and $O(n \log^{+} T)$ in the static and dynamic settings, respectively, for \RLS and \EA.
\end{proof}






\section{Reoptimization Times on Planar Graphs}
\label{sec4}

We also consider planar graphs with degree bound $\Delta \le 6$. It is well-known that all planar graphs can be colored with 4 colors, but the proof is famously non-trivial. Coloring planar graphs with 5 colors has a much simpler proof, and this setting was studied in~\cite{Sudholt2010b}. The reason for the degree bound ${\Delta \le 6}$ is that in~\cite{Sudholt2010b} it was shown that for every natural number~$c$ there exist tree-like graphs and a coloring where the ``root'' is $c$\nobreakdash-colored, and no Kempe chain or color elimination can improve this coloring. In the following we only consider the unbounded palette as no results for general planar graphs are known for bounded palette sizes.

\begin{theorem}
\label{thm:planar}
Consider adding up to $T$ edges to a 5-colored graph such that the resulting graph is planar with maximum degree $\Delta \le 6$.
Then the worst-case expected time for ILS with Kempe chains or  color eliminations to rediscover a proper 5-coloring is $O(n \log^+ T)$.
\end{theorem}
\begin{proof}
Lemma~\ref{lem:number-of-increased-colors} implies that, after inserting up to $T$ edges and running Grundy local search, at most $T$ vertices are colored 6 or 7. Lemma~\ref{lem:number-of-Delta-and-Delta-plus-1-non-increasing} showed that the number of vertices colored 6 or 7 is non-increasing.

In~\cite{Sudholt2010b} it was shown that for each vertex $v$ colored 6 or 7, there is a Kempe chain operation affecting a neighbour of~$v$ such that a color~$c$ at~$v$ becomes a free color and $v$ receives a color at most~5 after the next Grundy local search. If there are $i$ nodes colored 6 or 7, the probability of a Kempe chain move reducing the number of vertices colored with the highest color is at least $\Omega(i/n)$.

The same holds for color eliminations as in the aforementioned scenario, color~$c$ can be eliminated by a single Kempe chain. If there were other $c$-colored neighbors of $v$ not affected by the Kempe chain, this would be impossible. This means that a color elimination with the right parameters simulates the desired Kempe chain operation.

There are at most $T$ 7-colored nodes initially, and the expected time to recolor them is $O(n \log^+ T)$. Then there are at most~$T$ 6-colored nodes, and the same arguments yield another term of $O(n \log^+ T)$.
\end{proof}

\section{Faster Reoptimization Times Through \Aware Algorithms}
\label{sec5}

We now consider the performance of the original algorithms, but enhancing them with tailored operators that focus on the region of the graph that has been changed. The assumption for bounded-size palettes is that the algorithms are able to identify which edges are conflicting. This means that we are considering a gray box optimization scenario instead of a pure black-box setting. Since many of the previous results indicated that algorithm spend most of their time just finding the right vertex to apply mutation to, we expect the reoptimization times to decrease when using \aware operators.


The \EA and RLS are modified so that they mutate vertices that are part of a conflict with higher probability than other non-conflicting vertices. For the \EA we use a mutation probability of $1/2$ for the former and the standard mutation rate of $1/n$ for the latter. This is similar to fixed-parameter tractable evolutionary algorithms presented in~\cite{DBLP:journals/algorithmica/KratschN13} for the minimum vertex cover problem. Furthermore, step size adaptation which allows different amounts of changes per component of a given problem have been investigated for the dual formulation of the minimum vertex cover problem~\cite{DBLP:conf/foga/Pourhassan0N17}.


\begin{algorithm}[htb]
    \caption{\Aware \EA ($x$)}
    \algsetup{indent=1.5em}
    \begin{algorithmic}[1]
        \WHILE{optimum not found}

        \STATE Generate $y$ by deciding to mutate each $x_w$ with probability $1/2$ if $w$ is part of a conflict, and with probability $1/n$ otherwise. If yes, choose a new value $y_w \in \{1, \dots, k\} \setminus \{x_w\}$ uniformly at random. 
        \STATE If $y$ has no more conflicts than $x$, let $x := y$.
        \ENDWHILE
    \end{algorithmic}
    \label{alg:multi-aware-oneplusone}
\end{algorithm}

For RLS, the algorithm either flips a uniform random vertex that is part of a conflict or a vertex chosen uniformly at random from all vertices. The decision which strategy is used is made uniformly as well.

\begin{algorithm}[htb]
    \caption{\Aware RLS ($x$)}
    \algsetup{indent=1.5em}
    \begin{algorithmic}[1]
        \WHILE{optimum not found}
        \STATE Generate $y$ by choosing a vertex $w$ as follows. With probability $1/2$ choose $w$ uniformly at random from all vertices that are part of a conflict, otherwise choose $w$ uniformly at random from all vertices. Choose a new value $y_w \in \{1, \dots, k\} \setminus \{x_w\}$ uniformly at random and set $y_j = x_j$ for all $j \neq w$.\!\!\!
        \STATE If $y$ has no more conflicts than $x$, let $x := y$.
        \ENDWHILE
    \end{algorithmic}
    \label{alg:multi-aware-rls}
\end{algorithm}



For unbounded-size palettes, new edges can lead to higher color values emerging.
We work under the assumption that the algorithm is able to identify the vertices with the currently largest color. The \aware ILS algorithm then applies mutation to a vertex~$v$ chosen uniformly from all vertices with the largest color as follows. Color eliminations are applied to~$v$ directly. Kempe chains are most usefully applied in the neighborhood of~$v$, hence a neighbor of~$v$ is chosen uniformly at random.

\begin{algorithm}[hbt]
    \caption{\Aware ILS ($x$) with color eliminations (resp.\ Kempe chains)}
    \algsetup{indent=1.5em}
    \begin{algorithmic}[1]
        \STATE Replace $x$ by the result of Grundy local search applied to~$x$.
        \LOOP
            \STATE Let $w$ be a vertex chosen uniformly at random from all vertices with the largest color.
            \STATE Apply a color elimination to~$w$ (resp. apply a Kempe chain to a vertex chosen uniformly at random from the neighbors of~$w$) to generate a coloring~$y$.
            \STATE Let $z$ be the outcome of Grundy Local Search applied to~$y$.
            \STATE If $z \succeq x$ then $x := z$.
        \ENDLOOP
    \end{algorithmic}
    \label{alg:mutli-tailored-mutation-feasible}
\end{algorithm}

We argue that these \aware algorithms can be implemented efficiently, as stated in the following theorem.
\begin{theorem}
\label{the:execution-times-aware}
Consider the \aware RLS, the \aware \EA and the \aware ILS on a connected graph $G=(V, E)$ with $|V| \ge 2$ and maximum degree~$\Delta$.
If $b$ denotes the number of vertices currently involved in a conflict,
\begin{enumerate}
    \item one iteration of tailored RLS can be executed in
    expected time $O(\Delta)$,
    \item one iteration of the tailored \EA can be executed in expected time $O(\min\{b \Delta, \,  {|E|}\})$, and
    \item one iteration of tailored ILS with Kempe chains or color eliminations can be executed in time $O(|E|)$.
\end{enumerate}
\end{theorem}
Again, a proof is deferred to Appendix \ref{app:some_proofs} to keep the paper streamlined. Note that, in contrast to the bounds from Theorem~\ref{the:execution-times-unaware}, the bounds for the execution time of ILS are unchanged. The bound for RLS is now based on the maximum degree $\Delta$ instead of the average degree $2|E|/|V|$ since vertices that are part of a conflict may have an above-average degree. For graphs with $\Delta = O(|E|/|V|)$, e.\,g., regular graphs or graphs with $\Delta = O(1)$, both bounds are equivalent. For the \EA we get a much larger bound that is linear in the number of vertices that are part of a conflict, and never worse than $O(|E|)$. This is because all such vertices are mutated with probability $1/2$ and so determining the fitness of the offspring takes more time. It is plausible that the number of vertices that are part of a conflict quickly decreases during an early stage of a run, thus limiting these detrimental effects.

Revisiting previous analyses shows that in many cases the \aware algorithms have better runtime guarantees.
\begin{theorem}
\label{the:multi-aware-path-ea-rls}
If adding up to $T$ edges completes an $n$-vertex path, 
then the expected time to rediscover a proper 2-coloring is $O(n^2)$ for the \aware \EA and $O(n^2\log^+ T)$ for the \aware RLS.
%
\end{theorem}

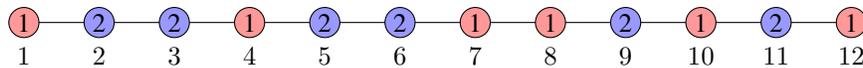
\begin{figure}[tb]
    \centering
    \begin{tikzpicture}
        \tikzstyle{vertex}=[draw,circle,inner sep=0cm,minimum size=.4cm];
        \tikzstyle{edge}=[black];

        \path (1, 0) edge [edge] ++(11, 0);

        \foreach \i/\c in {1/1, 2/2, 3/2, 4/1, 5/2, 6/2, 7/1, 8/1, 9/2, 10/1, 11/2, 12/1} {
            \node [vertex,color\c] (node\i) at (\i, 0) {\c};
            \node at (\i, -0.45) {$\i$};
        }
    \end{tikzpicture}
\caption{A colored path with vertices $\{1, \dots, 12\}$. When removing conflicting edges, the graph is decomposed into properly colored paths with vertex sets $\{1, 2\}, \{3, 4, 5\}, \{6, 7\}, \{8, 9, 10, 11, 12\}$.}
    \label{fig:paths-with-properly-colored-subpaths}
\end{figure}

\begin{proof}
Suppose there are $j \le T$ conflicting edges in the current coloring. Note that, when removing all conflicting edges, the graph decomposes into properly colored paths (see Figure~\ref{fig:paths-with-properly-colored-subpaths}). The vertex sets of these sub-paths form a partition of the graph's vertices.
%
By the pigeon-hole principle, the shortest of these properly colored paths has
length at most $n/j$.

These properly colored sub-paths can increase or decrease in length. For instance, the path $\{6, 7\}$ in Figure~\ref{fig:paths-with-properly-colored-subpaths} is shortened by 1 when flipping only vertex~6 or flipping only vertex~7. It is lengthened by~1 if only vertex~5 is flipped or only vertex~8 is flipped.
By the same arguments as in~\cite{Fischer2005}, recapped in the proof of Theorem~\ref{theorem:path}, the expected number of relevant steps to decrease the number of conflicts is $O(n^2/j^2)$ since we have a fair random walk on states up to $n/j$.

For the \aware \EA, the probability for a relevant step is $1/2$ as in each generation, a conflicting vertex in a shortest properly colored path is mutated with probability $1/2$. This results in an expected time bound of $O(n^2/j^2)$  for decreasing the number of conflicts from~$j$. Therefore, the worst-case expected time is at most $\sum_{j=1}^T O(n^2/j^2) =  O(n^2)$.

For \RLS, note that the only difference from the above analysis is that the probability for a relevant step now becomes $1/(2j)$. This results in a worst-case expected time bound of at most $\sum_{j=1}^T O(j \cdot (n^2/j^2))= O(n^2\log^+ T)$.
\end{proof}

Now we analyze the \aware algorithms with multiple conflicts for the depth-$2$ star. For RLS and both ILS algorithms we obtain an upper bound of $O(T)$, which is best possible in general as these algorithms only make one local change (modulo flipping the root of a depth-2 star) and $\Omega(T)$ local changes are needed to repair different parts of the graph. The \aware \EA only needs time $O(\log^+ T)$ as it can fix many conflicts in one generation.

\begin{theorem}
\label{the:multi-aware-rls-on-worst-case-tree-static}
If adding up to $T$ edges completes a depth-2 star, then the expected time to rediscover a proper 2-coloring is $O(T)$ for the \aware RLS and $O(\log^+ T)$ for the \aware \EA.
\end{theorem}
\begin{proof}
Let $C_t \le T$ denote the number of conflicts at time~$t$. For RLS, every vertex involved in a conflict is mutated with probability $1/(2C_t) + 1/(2n)$, which is at least $1/(2C_t)$ and at most $1/C_t$ as $C_t \le n$.
We show that the expected time to halve the number of conflicts, starting from $C_t$ conflicts, is at most $c \cdot C_t$ for some constant~$c > 0$. For all $t' \ge t$, as long as $C_{t'} > C_t/2$, the probability of mutating a vertex involved in a conflict is at least $1/(2C_t)$ and at most $2/C_t$.

Consider a conflict on a path $P_i$ from the root to a leaf. This conflict can be resolved
as argued in the proof of Theorem~\ref{the:rls-on-worst-case-tree-static}. If both edges of $P_i$ are conflicting, flipping the middle vertex (and not flipping any other vertices of $P_i$) resolves both conflicts. Otherwise, if the conflict involves a leaf node, flipping said leaf and not flipping any other vertices of $P_i$ resolves all conflicts on $P_i$. Finally, if the conflict involves the edge at the root, it can be resolved by first flipping the middle vertex and then flipping the leaf, and not flipping any other vertices of $P_i$ during these steps.

In all the above cases, a lower bound on the probability of resolving all conflicts on the path $P_i$ during a phase of $2C_t$ generations, or decreasing the number of conflicts to a value at most $C_t/2$, is at least
\[
\binom{2C_t}{2} \cdot \frac{1}{2C_t} \cdot \frac{1}{2C_t} \left(1-\frac{2}{C_t}\right)^{2(2C_t-2)}, \]
which is bounded from below by a positive constant for $C_t \ge 3$.
The term ${(1-2/C_t)^{2(2C_t-2)}}$ reflects the probability of the event that up to 2 specified vertices do not flip during $2C_t-2$ iterations.
Note that these products $(1-2/C_t)$ can be dropped for iterations in which the number of conflicts has decreased to $C_t/2$ (and the upper probability bound of $2/C_t$ might not hold).
Also note that the above events are conditionally independent for all paths $P_i$ that have conflicts, assuming that the root does not flip. There are at least $C_t/2$ such paths at the start of the period of $2C_t$ generations. Hence, the expected number of conflicts resolved in a period of $2C_t$ generations is at least $c \cdot C_t$, for a constant~$c > 0$, or the number of conflicts has decreased to a value at most $C_t/2$.
By additive drift, the expected time for the number of conflicts to decrease to $C_t/2$ or below is at most $C_t/c$.

This implies that the expected time for $C_t$ to decrease below $3$ is at most
\[
\frac{T}{c} + \frac{T}{2c} + \frac{T}{4c} + \dots \le \frac{2T}{c}.
\]
The expected time to resolve the final at most 2 conflicts is $O(1)$ by considering the same events as above.

For the \EA, we note that in any consecutive two generations, conditioned on the event that the root does not flip, a conflict in path $P_i$ gets resolved with probability at least
\[\frac12\cdot \left(1-\frac{1}{n}\right)\cdot \frac12\cdot \frac12\geq \frac{1}{16},\]
as in the first generation, with probability $\frac12(1-\frac{1}{n})$, the middle vertex is flipped and the leaf is not flipped, and in the second generation, with probability $\frac12\cdot \frac12$, the leaf is flipped and the middle vertex is not flipped.

Thus, by the linearity of expectation and the fact that, with probability at least $1/4$, the root is not flipped in two generations, we know that
\[
E[C_{t+2}] \leq C_t \left(1-\frac{1}{16}\cdot \frac14 \right) = \frac{63}{64} \cdot C_t,
\]
where $C_t$ is the number of conflicting paths at time $t$.
Therefore, by the multiplicative drift theorem~\cite{Doerr2012}, the expected time to reduce the number of conflicts from at most~$T$ to $0$ is $O(\log^+ T)$.
\end{proof}

\begin{theorem}
\label{the:aware-T-path-tree-Kempe}
Consider a dynamic graph that is a path or binary tree after inserting $T$ edges.
The expected time for \aware ILS with Kempe chains to rediscover a proper 2-coloring is $O(T)$.
\end{theorem}
\begin{proof}
For paths, the largest color that can emerge through added conflicting edges and the following Grundy local search is~3. \Aware ILS picks a random 3-colored vertex~$v$ and applies either a color elimination to~$v$ or a Kempe chain to a neighbor of~$v$. In both cases, choosing appropriate colors will create a free color for~$v$ and the number of 3-colored vertices decreases. Since the probability of choosing appropriate colors is $\Omega(1)$, the expected time to reduce the number of 3-colored vertices is $O(1)$. Since this has to happen at most~$T$ times, an upper bound of $O(T)$ follows.

For binary trees, color values of~$4$ can emerge during Grundy local search (but no larger color values since the maximum degree is~3). By Lemma~\ref{lem:number-of-Delta-and-Delta-plus-1-non-increasing}, the number of vertices colored 3 or 4 cannot increase.
As argued above for paths, the expected time until color 4 disappears is $O(T)$. By then, there are at most $T$ 3-colored vertices and the time until these disappear is $O(T)$ by the same arguments.
\end{proof}

\begin{theorem}
\label{the:aware-T-bipartite-color-elimination}
Consider a dynamic graph that is bipartite after inserting $T$ edges.
Then \aware ILS with color eliminations re-discovers a proper 2\nobreakdash-coloring in $O(\min\{\sqrt{T}, \Gamma\} n)$ iterations where $\Gamma$ is the Grundy-number after inserting the edges.

If no end point of an added edge is neighbored to an end point of another added edge, or if $\Gamma \le 4$, \aware ILS with color eliminations re-discovers a proper 2-coloring in $O(T)$ expected iterations.
\end{theorem}
\begin{proof}
The proof follows from the proof of Theorem~\ref{the:ils-ce-rediscover-time} and that the maximum color is bounded by $\min\{\sqrt{T}, \Gamma\}$. The expected time to eliminate the largest color is at most~$n$: there are at most $n$ vertices with the largest color. In every iteration, the algorithm applies color eliminations to a vertex~$v$ of the largest color, and every color elimination creates a free color that allows~$v$ to receive a smaller color in the Grundy local search. (The time bound is $n$ instead of $T$ since, as mentioned in the proof of Theorem~\ref{the:ils-ce-rediscover-time} the number of vertices with large colors can increase if the number of vertices with the largest color decreases.)

If no end point of an added edge is neighbored to an end point of another added edge, or if $\Gamma \le 4$, then the largest color is at most~4 and the time to eliminate at most $T$ occurrences of color~4 and at most $T$ occurrences of color~3 is~$O(T)$.
\end{proof}

\begin{theorem}
\label{the:aware-T-planar}
Consider adding $T$ edges to a 5-colored graph such that the resulting graph is planar with maximum degree $\Delta \le 6$.
Then the worst-case expected time for \aware ILS with Kempe chains or color eliminations to rediscover a proper 5-coloring is $O(T)$.
\end{theorem}
\begin{proof}
This result follows as in the proof of Theorem~\ref{thm:planar}. The only difference is that every mutation only affects vertices of the currently largest color (color eliminations) or neighbors thereof (for Kempe chains). The proof of Theorem~\ref{thm:planar} has shown that the probability of a mutation being improving is $\Omega(1)$. Hence, the probability of reducing the number of 7-colored vertices is $\Omega(1)$ and in expected time $O(T)$, all 7-colored vertices are eliminated. Since, as shown in the proof of Theorem~\ref{thm:planar}, there are at most $T$ 6-colored vertices, the same arguments apply to the number of 6-colored vertices.
\end{proof}

Despite these positive results for \aware operators, they cannot prevent exponential times as shown for binary trees and depth-2 stars.
\begin{theorem}
\label{theorem:aware-lower-tree}
If adding an edge completes an $n$-vertex complete binary tree, the worst-case expected time for the \aware \EA to rediscover a proper 2-coloring is $\Omega\big(n^{(n-3)/4}\big)$. The \aware RLS is unable to rediscover a proper 2-coloring in the worst case.
\end{theorem}
\begin{proof}
The proof is similar to proof of Theorem \ref{theorem:tree}. The Hamming distance between any worst-case coloring in the set $A_0$ and any other acceptable coloring is still at least $\frac{n+1}{4}$. We can save a factor of $n$ as the algorithm will mutate each of the endpoints of the conflict edge $(u,v)$ with $1/2$ probability, rather than with probability $1/n$ as before.
\end{proof}

\begin{theorem}
\label{the:aware-worst-case-tree}
On the depth-2 star from Theorem~\ref{the:worst-case-for-Kempe-chains}, \aware ILS with Kempe chains needs $\Theta(2^{n/2})$ generations in expectation to rediscover a proper 2\nobreakdash-coloring.
\end{theorem}
\begin{proof}
\Aware ILS with Kempe chains applies a Kempe chain to uniformly chosen neighbors of the root. The transition probabilities still follow an Ehrenfest urn model; the only difference is that no Kempe chain can originate from the root itself. This does not affect the proof of Theorem~\ref{the:worst-case-for-Kempe-chains}, and the same result applies.
\end{proof}

\section{Discussion and Conclusions}

We have studied graph vertex coloring in a dynamic setting where up to $T$ edges are added to a properly colored graph. We ask for the time to re-discover a proper coloring based on the proper coloring of the graph prior to the edge insertion operation. Our results in Table~\ref{tab:all-times} show that reoptimization can be much more efficient than optimizing from scratch, i.e., neglecting the existing proper coloring. In many upper bounds a factor of $\log n$ can be replaced by $\log^+ T = \max\{1, \log T\}$ and we showed a tighter general bound for bipartite graphs of $O(\min\{\sqrt{T}, \Gamma\} n \log n)$ as opposed to $O(n^2 \log n)$~\cite{Sudholt2010b}.
However, this heavily depends on the graph class and algorithms. For instance, depth-2 stars led to exponential times for Kempe chains and times of $O(n \log^+ T)$ for all other algorithms.
Reoptimization can also be more difficult as we can naturally create worst-case initial colorings which are very unlikely in the static setting. On paths and binary trees the dynamic setting allows for negative results that are stronger than those previously published~\cite{Fischer2005,Sudholt2005}.

\Aware operators put a higher probability on mutating vertices involved in conflicts (for bounded-size palettes) or that have large colors (for unbounded-size palettes). This improves many upper bounds from $O(n \log^+ T)$ to $O(T)$. For the \EA on depth-2 stars the expected time even decreases to $O(\log^+ T)$. However, tailored algorithms cannot prevent inefficient runtimes in settings where the corresponding generic algorithm is inefficient.

Our analyses concerned the number of iterations. When considering the execution time as the number of elementary operations (see Theorems~\ref{the:execution-times-unaware} and~\ref{the:execution-times-aware}), on planar graphs with $\Delta \le 6$ ILS rediscovers a proper 5-coloring in expected $O(T)$ iterations. This translates to $O(nT)$ elementary operations using Theorem~\ref{the:execution-times-aware} and the fact that for planar graphs $G=(V, E)$ we have $|E| = O(|V|)$. This is generally faster than the $O(n^2)$ bound for the problem-specific algorithm from~\cite{Robertson1996} that solves the static problem. The latter algorithm guarantees a 4-coloring though, whereas we can only guarantee a 5-coloring.\footnote{Given the complexity of the proof of the famous Four Color Theorem and the algorithm from~\cite{Robertson1996}, we would not have expected a simple proof that guarantees a 4-coloring.}
For dynamic coloring algorithms, as mentioned in Section~\ref{sec:related-work}, there is a forest, which is a planar graph, on which dynamically maintaining a $c$-coloring requires recoloring at least $\Omega(n^{2/c(c-1)})$ vertices per update on average, for any $c\geq 2$. Setting $c=5$ yields a lower bound of $\Omega(n^{1/10})$
for
rediscovering
a $5$-coloring of the mentioned planar graph.


For future work, it would be interesting to study the \unaware and \aware vertex coloring algorithms on broader classes of graphs. Furthermore, the performance of evolutionary algorithms for other graph problems (e.g., maximum independent set, edge coloring) is largely open.

\paragraph{Acknowledgements} J.~Bossek acknowledges support by the \href{https://www.ercis.org}{\textit{European Research Center for Information Systems (ERCIS)}}.
F.~Neumann has been supported by the Australian Research Council (ARC) through grant DP160102401.

%
%

\bibliographystyle{unsrt}
\bibliography{arxiv_v1.bib}

%
%


\appendix

\section{Implementation Notes and Analysis of Execution Times}\label{app:some_proofs}

Here we present proofs of Theorems~\ref{the:execution-times-unaware} and Theorem~\ref{the:execution-times-aware} on the computational complexity of executing one iteration of the considered algorithms.
\begin{proof}[Proof of Theorem~\ref{the:execution-times-unaware}]
We store the graph $G$ as an adjacency list, with an array of vertices and every vertex storing a linked list or array of all its neighbors. We further assume that all vertices have space to store their current color as well as temporary markers for graph traversals, and for temporarily marking vertices to be recolored.

Since all vertices are stored in an array, we can choose a vertex uniformly at random in time $O(1)$.
Consequently, the mutation step in RLS takes time $O(1)$. We still need to consider the selection step, though. A naive implementation might copy the parent, then perform mutation and then compute the number of conflicts in the mutant from scratch, in a graph traversal that takes time $\Theta(|V|+|E|)$. With a more clever implementation, we can be much faster and avoid some of these steps. Let $v$ be the vertex selected for mutation and let~$i$ be the new color chosen for~$v$ in the offspring, then we can simply check all neighbors of~$v$ and count how many neighbors have the same color as~$v$ and how many neighbors have color~$i$. The difference of these quantities determines whether recoloring~$v$ would decrease the fitness or not. If it does, the generation is complete. Otherwise, we do recolor~$v$ with color~$i$.
This can be done in time $O(1) + c\deg(v)$, when $v$ is fixed, for a suitable constant~$c > 0$. Since $v$ is chosen uniformly at random, we get an upper bound of
\begin{equation}
    O(1) + \frac{1}{|V|} \sum_{v \in V} c\deg(v) = O(1) + \frac{2c|E|}{|V|} = O(|E|/|V|)
    \label{eq:edges-by-vertices}
\end{equation}
since $\sum_{v \in V} \deg(V) = 2|E|$ (and $|E| \ge |V|-1 \ge 1$ implies $O(1) \subseteq O(|E|/|V|)$).


For the \EA we can proceed in a similar way. Instead of deciding individually for each vertex whether it should be recolored or not (which would take time $\Theta(|V|)$), we first compute the number of vertices to be recolored according to the distribution $\Prob{\text{recolor $i$ vertices}} = \binom{|V|}{i} (1/|V|)^{i} (1-1/|V|)^{|V|-i}$. Then we uniformly select $i$ different vertices to be recolored and proceed as for RLS (taking care to correctly count edges for which both endpoints are to be recolored; this can be done using temporary markers for these vertices). Since the expected number of recolored vertices is $O(1)$, the expected time to execute one iteration of the (1+1)~EA, including expected time $O(1)$ for setting and clearing markers, is $O(1) + O(1) \cdot O(|E|/|V|) = O(|E|/|V|)$.



A Kempe chain can be implemented in time $O(|V|+|E|)$ with a graph traversal on the subgraph $H_j(v)$. For instance, we can run a depth-first search (DFS) that only considers neighbors colored~$i$ or~$j$; this makes DFS run on the subgraph $H_j(v)$. A color elimination also runs in time $O(|V|+|E|)$ as it concerns a Kempe chain on a union of subgraphs. More specifically, if $v_1, \dots, v_{\ell}$ denote all $i$-colored neighbors of vertex~$v$, these vertices may belong to different subgraphs $H_j(v_\cdot)$ (see Section~\ref{sec:algorithms}) or multiple neighbors might be part of the same subgraph. To account for this, we can start DFS at $v_1, \dots, v_\ell$ in any given order and skip vertices that have already been visited in a previous DFS call. This may require the use of markers that can be set and cleared in time $O(|V|)$.

We still need to account for the time to execute Grundy local search and selection.
As shown in~\cite{HedJacSri03}, Grundy local search runs in time $O(|V|+|E|)$. Selection is based on the number of conflicts and the color-occurrence vector (see Section~\ref{sec:algorithms}). After every run of Grundy local search, the coloring is feasible and then selection is purely based on the color-occurrence vector (unless a dynamic change occurs). The color-occurrence vector of the offspring can either be computed from scratch or be computed incrementally from that of the parent by updating the color counters during mutation and local search. In both cases, the additional time for this is bounded by $O(|V|+|E|)$. Hence the total time to execute one generation of ILS with either mutation operator is $O(|V|+|E|)$. Since the graph is connected and $|V| \ge 2$ implies $|E| \ge 1$, we have $O(|V|+|E|) = O(|E|)$.
\end{proof}

Now we analyse the execution times of \aware algorithms and give a proof of Theorem~\ref{the:execution-times-aware}.
\begin{proof}[Proof of Theorem~\ref{the:execution-times-aware}]
For the \aware RLS we only need to consider the case that the algorithm decides to select a vertex~$w$ is chosen uniformly at random from all vertices that are part of a conflict. To implement this efficiently,
we use an idea from~\cite{henzinger2020constant}: we maintain flags for each vertex that indicate whether the vertex is currently part of a conflict as well as a separate array $A = A[1] \dots A[|V|]$ of size $|V|$ that stores all vertices that carry a positive flag. In addition, every vertex with a positive flag stores its position in the array $A$. We maintain a value $s$ that reflects the number of such vertices currently present in the array. Then picking a vertex from this array uniformly at random in time $O(1)$ is straightforward: pick an index $i$ uniformly at random from $\{1, \dots, s\}$ and return the vertex stored at $A[i]$.

We argue that the array can be maintained efficiently. When a vertex $w$ is being recolored following a positive selection, we check $w$ and its neighbors. If a vertex~$v$ becomes part of a conflict, the flag is set, $s$ is incremented and $v$ is inserted into the array as $A[s]$. We store the position $s$ in the vertex~$v$. If a vertex~$v$ is no longer part of a conflict, we check $v$'s position in the array. Let this be~$i$. Then $v$'s flag is reset, element $A[s]$ is copied to $A[i]$ and $s$ is decremented. This way, the vertex is removed from the array in time $O(1)$.
The time for supporting an update (i.e., adding or removing a vertex with a flag) to the data structure and for sampling a uniform vertex from $A$ is $O(1)$. Since this is done for $w$ and its neighbors, the total effort for one iteration of the \aware RLS is $O(\Delta)$.

Note that the previous estimation~\eqref{eq:edges-by-vertices} from the proof of Theorem~\ref{the:execution-times-unaware} that lead to the bound $O(|E|/|V|)$ no longer applies if we only select from vertices with a conflict as the average degree of these vertices may be larger than $\Theta(|E|/|V|)$.
However, we can argue that, by the same argument as in the proof of Theorem \ref{the:execution-times-unaware}, the time for recoloring a conflicting vertex $v$ is $O(1)+c\deg(v)=O(\Delta)$, where $\Delta$ is the maximum degree of the graph.

Note that for the \EA the effort for executing an iteration may increase, compared to RLS, because of the higher mutation rate of $1/2$ for vertices that are part of a conflict. When there are $b$ vertices $v_1, \dots, v_b$ being part of a conflict, an iteration can still be executed in time $O(1) \cdot \sum_{i=1}^b \deg(v_i)$. This is bounded by $O(b \Delta)$. The running time $O(\min\{b\Delta, |E|\})$ for the tailored \EA then follows from the observation that an iteration can always finish in time $O(|V|+|E|) = O(|E|)$.

For \aware ILS we need to be able to choose from vertices with the largest color. To implement this efficiently, we may use arrays $A_1, A_2, \dots, A_{\Delta+1}$ such that $A_i$ stores all vertices that are currently colored~$i$. These arrays can be set up and maintained as described above for RLS and the array~$A$. The largest color $c$ can clearly be determined in time $O(|V|)$ and picking a uniform random vertex from $A_c$ takes time $O(1)$. Thus, the computational complexity only increases by a constant factor and the previous bound of $O(|E|)$ still applies (recall that $|V| = O(|E|)$).
\end{proof}

\end{document}